\documentclass{article}

\usepackage[preprint]{neurips_2026}

\usepackage[utf8]{inputenc} % allow utf-8 input
\usepackage[T1]{fontenc}    % use 8-bit T1 fonts
\usepackage{hyperref}       % hyperlinks
\usepackage{url}            % simple URL typesetting
\usepackage{booktabs}       % professional-quality tables
\usepackage{amsfonts}       % blackboard math symbols
\usepackage{nicefrac}       % compact symbols for 1/2, etc.
\usepackage{microtype}      % microtypography
\usepackage{xcolor}         % colors
\usepackage{amsmath}
\usepackage{graphicx}
\usepackage{algorithm}
\usepackage{amsmath}
\usepackage{amssymb}
\usepackage{mathtools}
\usepackage{multirow}
\usepackage{comment}
\usepackage{algpseudocode}
\usepackage{enumitem}

\usepackage{booktabs} 
\usepackage{tikz}
\usetikzlibrary{arrows.meta, positioning, fit, shapes.geometric}
\usepackage{url}
\usepackage{hyperref}
\usepackage{xcolor}
\usepackage{mdframed}

\usepackage{listings}

\usepackage[most]{tcolorbox}
\usepackage{fvextra}
\usepackage{xcolor}
\newcommand{\paradigm}[1]{\textit{#1}}
\definecolor{promptbg}{RGB}{248,248,248}
\definecolor{promptframe}{RGB}{210,210,210}

\DefineVerbatimEnvironment{PromptVerbatim}{Verbatim}{
  breaklines=true,
  breakanywhere=false,
  fontsize=\footnotesize,
  fontfamily=tt,
  tabsize=2,
  obeytabs=true
}

\newenvironment{promptbox}
  {%
    \VerbatimEnvironment
    \begin{tcolorbox}[
      enhanced,
      breakable,
      colback=promptbg,
      colframe=promptframe,
      boxrule=0.4pt,
      arc=2pt,
      left=6pt,
      right=6pt,
      top=6pt,
      bottom=6pt
    ]
    \begin{PromptVerbatim}
  }
  {%
    \end{PromptVerbatim}
    \end{tcolorbox}
  }

\lstdefinestyle{pyinline}{
     language=Python,
     basicstyle=\ttfamily\footnotesize,
     keywordstyle=\color{black!70}\bfseries,
     commentstyle=\color{gray}\itshape,
     breaklines=true,
     showstringspaces=false,
   }

\title{%Are Large Language Models Good Heuristic Generators for Combinatorial Search?
Formalize, Don't Optimize: The \emph{Heuristic Trap} in LLM-Generated Combinatorial Solvers

}

\author{%
  \textbf{Haoyu Wang}\textsuperscript{1} \quad
  \textbf{Yuliang Song}\textsuperscript{2} \quad
  \textbf{Tao Li}\textsuperscript{3} \quad
  \textbf{Zhiwei Deng}\textsuperscript{3} \\
  \textbf{Yaqing Wang}\textsuperscript{3} \quad
  \textbf{Deepak Ramachandran}\textsuperscript{3} \quad
  \textbf{Eldan Cohen}\textsuperscript{2} \quad
  \textbf{Dan Roth}\textsuperscript{1,4} \\[4pt]
  \normalfont
  \textsuperscript{1}University of Pennsylvania \quad
  \textsuperscript{2}University of Toronto \quad
  \textsuperscript{3}Google DeepMind \quad
  \textsuperscript{4}Oracle AI \\[4pt]
  \texttt{\{why16gzl, danroth\}@seas.upenn.edu} \quad
  \texttt{\{yl.song, eldan.cohen\}@utoronto.ca} \\
  \texttt{\{tlinlp, zhiweideng, yaqingwang, ramachandrand\}@google.com}
}

\begin{document}

\maketitle
\begin{abstract}
    Large Language Models (LLMs) struggle to solve complex combinatorial problems through direct reasoning, so recent neuro-symbolic systems increasingly use them to synthesize executable solvers. A central design question is how the LLM should represent the solver, and whether it should also attempt to optimize search. We introduce CP-SynC-XL, a benchmark of $100$ combinatorial problems (4{,}577 instances), and evaluate three solver-construction paradigms: native algorithmic search (\paradigm{Python}), constraint modeling through a Python solver API (\paradigm{Python + OR-Tools}), and declarative constraint modeling (\paradigm{MiniZinc + OR-Tools}).
    We find a consistent \textit{representational divergence}: \paradigm{Python + OR-Tools} attains the highest correctness across LLMs, while \paradigm{MiniZinc + OR-Tools} has lower absolute coverage despite using the same OR-Tools back-end. Native \paradigm{Python} is the most likely to return a schema-valid solution that fails verification, whereas solver-backed paths preserve higher conditional fidelity. On the heuristic axis, prompting for search optimization yields only small median speed-ups ($1.03$--$1.12\times$) and a strongly bimodal effect: many instances slow down, and correctness drops sharply on a long tail of problems.
    A paired code-level audit traces these regressions to a recurring \textbf{heuristic trap}. Under an efficiency-oriented prompt, the LLM may replace complete search with local approximations (\paradigm{Python}), inject unverified bounds (\paradigm{Python + OR-Tools}), or add redundant declarative machinery that overwhelms or over-constrains the model (\paradigm{MiniZinc + OR-Tools}).
    These findings support a conservative design principle for LLM-generated combinatorial solvers: use the LLM primarily to formalize variables, constraints, and objectives for verified solvers, and separately check any LLM-authored search optimization before use.
\end{abstract}

\section{Introduction}
\label{sec:introduction}
Large Language Models (LLMs) have demonstrated strong performance on mathematical reasoning \citep{mirzadeh2024gsm, jaech2024openai, hilbert} and general-purpose code synthesis \citep{liu2024deepseek, comanici2025gemini}, yet complex combinatorial optimization remains a difficult setting. While LLMs can navigate basic logical puzzles \citep{shojaee2025illusion}, their reasoning degrades on the large search spaces and implicit domain constraints characteristic of real-world combinatorial problems. As problem complexity scales, the autoregressive generation process becomes a ``reasoning bottleneck''~\citep{fan-etal-2024-nphardeval, jiang2026reasoning}: without reliable internal state tracking for forward-checking and backtracking, models commit to early sub-optimal decisions and may lose global consistency across long-horizon constraints.

To mitigate this limitation, recent work augments LLMs with external control structures such as A* search \citep{zhuang2023toolchain}, Monte Carlo Tree Search (MCTS) \citep{liu-etal-2025-autoct}, or agentic working memory \citep{xu2025mem}. These interactive approaches explicitly maintain states and expose intermediate decisions to backtracking or search. However, using an LLM as an interactive search agent in NP-hard domains incurs high inference latency and token cost \citep{yang2024large,jiang2026reasoning}; moreover, external memory does not by itself remove heuristic blind spots that can steer the process toward local optima \citep{wang2025polynomialtime}.
These limitations motivate shifting the computational burden away from direct textual reasoning. One promising strategy leverages LLM-driven program synthesis \citep{duchnowski-etal-2025-knapsack, michailidis2025cp}: the LLM writes an instance-agnostic executable artefact that formalizes the problem requirements, while the actual combinatorial search either runs inside that artefact or is delegated to a verified backend.

\begin{figure}[t]
\centering
\includegraphics[width=\textwidth]{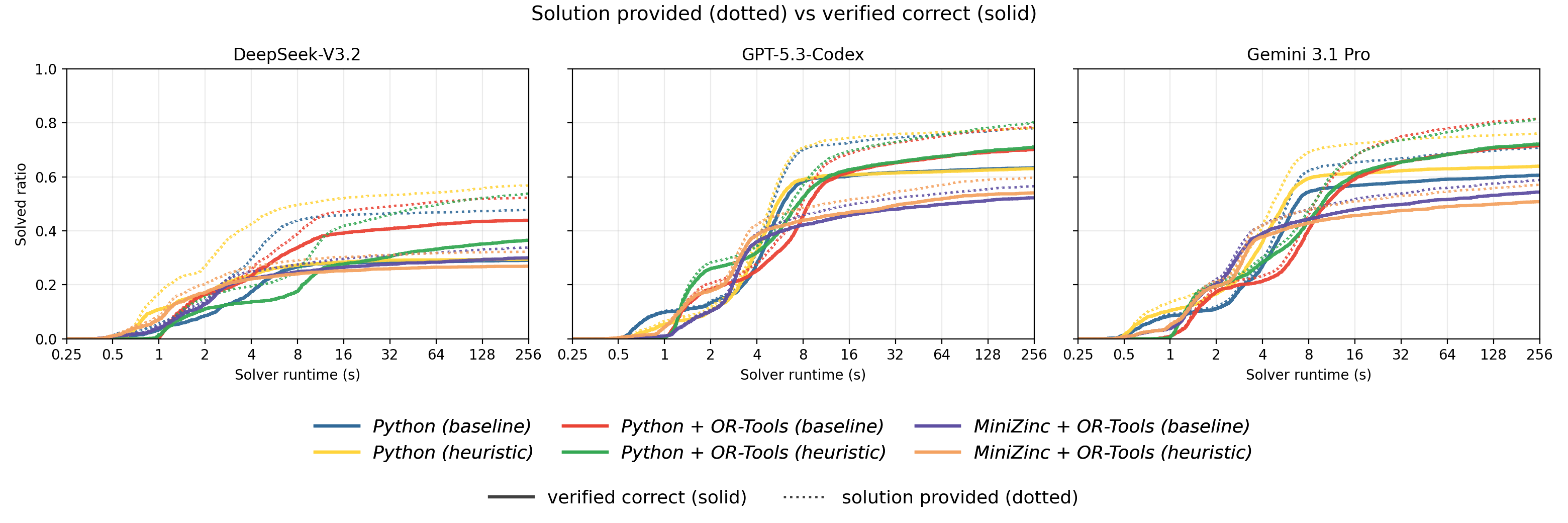}
\caption{Solution provided (dotted) vs.\ verified correct (solid) by paradigm and prompt, one panel per LLM (log time axis). \paradigm{Python + OR-Tools} attains the highest correctness at the $256$\,s tail on every LLM; the dotted--solid gap is largest on \paradigm{Python} and smallest on \paradigm{MiniZinc + OR-Tools}; the heuristic prompt fails to reliably dominate the baseline, previewing the \textbf{heuristic trap}.}
\label{fig:km}
\end{figure}

Two open questions run through this synthesis setting: \emph{how} should the LLM represent the solver, and \emph{whether} the LLM should actively optimize search by generating heuristics. 
On representation, we find a \textit{representational divergence}: LLMs are more successful with a Python solver API than with a declarative modeling language, even when both use the same OR-Tools back-end. On search optimization, we use \textbf{heuristic trap} to denote the recurring pattern in which an efficiency-oriented prompt induces unverified bounds, local approximations, or overcomplicated constraints that sometimes target speed but reduce correctness or fidelity.

We study these questions by evaluating three frontier LLMs (\textsc{GPT-5.3-Codex}, \textsc{Gemini 3.1 Pro}, \textsc{DeepSeek-V3.2}) on CP-SynC-XL, an extension of CP-SynC~\citep{song2026cpsync}, with $100$ combinatorial problems and $4{,}577$ instances. We compare three solver-construction paradigms: native algorithmic search (\paradigm{Python}), constraint modeling via a Python solver API (\paradigm{Python + OR-Tools}), and declarative constraint modeling (\paradigm{MiniZinc + OR-Tools}). The contrast we study is the representational surface, with a shared CP-SAT back-end for the two solver-backed paradigms. Figure~\ref{fig:km} previews both findings: \paradigm{Python + OR-Tools} attains the highest correctness on every LLM, while the heuristic prompt fails to reliably dominate the baseline and can regress sharply on the weakest LLM.
These experiments provide an instance-level view of when LLM-generated
combinatorial solvers benefit from symbolic delegation and when efficiency-oriented
prompting introduces additional risk. Our main contributions are:

\begin{enumerate} %[noitemsep]
 \item \textbf{CP-SynC-XL benchmark}: we introduce CP-SynC-XL, a benchmark containing 4,577 parameterized instances for 100 heterogeneous Constraint Satisfaction Problems and Constraint Optimization Problems for evaluating LLM-synthesised solvers.
 \item \textbf{Representational divergence}: we show that solver correctness depends primarily on the LLM's familiarity with the modeling surface rather than with the underlying solver---\paradigm{Python + OR-Tools} attains the highest correctness on every LLM despite sharing CP-SAT with \paradigm{MiniZinc + OR-Tools}, while native \paradigm{Python} exhibits the highest rate of silent failures.
 \item \textbf{Heuristic trap}: we show that prompting LLMs to optimize search is unreliable---a small median speed-up masks a sharply bimodal cost associated with six paradigm-specific failure modes that iterative refinement often preserves rather than repairs; our findings support keeping the LLM close to formalization rather than asking it to supply search optimizations.
\end{enumerate}

\section{Related Work}

\textbf{LLMs as direct combinatorial solvers.}
Recent benchmarks show that prompting LLMs to solve NP-hard problems
end-to-end deteriorates sharply with instance scale~\citep{fan-etal-2024-nphardeval, jiang2026reasoning},
and that this compositional limit is structural rather than purely
informational~\citep{dziri2023faith}. Performance is also sensitive to
problem representation~\citep{duchnowski-etal-2025-knapsack}: implicit
constraints and natural-language framings are systematically harder
than equivalent formal encodings. These findings motivate the shift
from direct textual reasoning to symbolic delegation.

\textbf{LLM-as-translator.}
A second line replaces direct reasoning with translation into a formal
language consumed by an external solver. Logic-LM~\citep{pan2023logic}
and SATLM~\citep{ye2023satlm} translate natural-language problems into
FOL/SAT for off-the-shelf provers, and PAL~\citep{gao2023pal}
pioneered the broader Program-Aided LM template. In OR-style modeling,
NL4Opt~\citep{ramamonjison2023nl4opt} and
OptiMUS~\citep{ahmaditeshnizi2024optimus} target MILP, while
\citet{haoyu2025llms} and \citet{song2025llmcp} target CSP/CP through
dedicated data structures and declarative MiniZinc and PyCSP3
models~\citep{nethercote2007minizinc, lecoutre2020pycsp3}. These
studies establish that LLMs can often translate problem descriptions
into solver-consumable artefacts, but they typically evaluate one
surface at a time, one or a few instances per problem, or problem-level
success rather than a paired instance-level outcome decomposition. As a
result, it is difficult to tell whether a reported success rate comes
from the underlying solver, the LLM's fluency with a particular
modeling surface, or the verifier's tolerance for
semantically incorrect solutions. We instead compare three
representational surfaces for the same problem set and LLMs, use
OR-Tools CP-SAT as the common back-end for the two solver-backed
conditions, and separate \emph{provided}, \emph{correct}, and
conditional fidelity at the instance level.

% removed: wrong-but-schema-valid output

\textbf{LLMs as algorithmic / heuristic designers.}
A more recent line uses LLMs to synthesize the algorithm itself.
FunSearch~\citep{romera2024mathematical} evolves heuristic programs
against a fitness function, Eureka~\citep{ma2023eureka} discovers
reward functions, and Evolution-of-Heuristics~\citep{liu2024evolution}
couples LLM proposal with evolutionary selection; in interactive
settings, \citet{wang2025polynomialtime} treat the LLM as a
polynomial-time heuristic queried inside MCTS. The reported successes
typically rely on a problem-specific scoring oracle, a
population-and-selection loop with many LLM samples, and the freedom to
specialize each heuristic to a single problem family. We adopt the
synthesis framing but study a regime closer to many deployment
settings: single-shot generation with a bounded refinement loop, a
uniform rubric across $100$ problems, and no test-time
evolution. 
%In this regime we use \emph{heuristic trap} to denote the
%recurring pattern in which an efficiency-oriented prompt induces
%unverified bounds, incomplete substitutions, or over-specified models
%that sometimes improve runtime but can reduce correctness.

% =====================================================================
% Method section for NeurIPS submission.
% =====================================================================

\section{Method}
\label{sec:method}

To investigate the two questions raised in Sec.~\ref{sec:introduction}---\emph{how} should the LLM represent the solver, and \emph{whether} it should attempt to optimize search---we use a controlled factorial design with two axes: the solver-generation \emph{paradigm} and the \emph{prompt}. For each natural-language combinatorial problem, the LLM produces a reusable artefact that consumes instance parameters at runtime and returns a solution in the benchmark's required format. The paradigm fixes the representational surface, spanning native algorithmic search (\paradigm{Python}), constraint modeling through the OR-Tools CP-SAT Python API (\paradigm{Python + OR-Tools}), and declarative modeling in MiniZinc (\paradigm{MiniZinc + OR-Tools}). The prompt toggles between a \textit{baseline} that asks only for correctness and a \textit{heuristic} variant that additionally asks for efficiency-oriented design. Within each comparison, we hold fixed the LLM, decoding parameters, instances, evaluator, timeouts, and refinement budget, so differences can be attributed to the paradigm/prompt factors rather than to incidental implementation choices.

\subsection{Solver-Generation Paradigms}
\label{sec:method-paradigms}

All paradigms run inside the same evaluation framework: a single
orchestration loop that prompts the LLM, executes the generated solver,
and validates its output. They share the same problem description,
input schema, and required output keys; the per-paradigm paragraphs
below describe what artefact the LLM is asked to produce in each case
and which solver backend it is allowed to use.

\textbf{Native algorithmic search (\paradigm{Python}).}
This paradigm tests the LLM as an algorithm designer. The LLM receives the problem context and is instructed to generate a standalone Python algorithm for solving the problem using only the Python standard library. At runtime, the generated function takes the input instance and returns a solution in the required format. %$P_{\mathrm{OUT}}$. 
We prohibit the use of backend solving tools, forcing the LLM to design the full algorithm itself through strategies such as pruning, dynamic programming, branch-and-bound, or problem-specific methods.

% The LLM emits a single Python function,
% \texttt{solve\_instance(}\allowbreak\texttt{data\_dict)}, using only the Python
% standard library. The function parses the instance parameters at
% runtime and returns a solution dictionary. With no optimization
% back-end available, the LLM is responsible for the full algorithm:
% search order, pruning, dynamic programming, branch-and-bound, or any
% other procedure it chooses to implement.
% This paradigm tests the LLM as an algorithm~designer.

\textbf{Constraint modeling via the OR-Tools Python API (\paradigm{Python + OR-Tools}).}
This paradigm tests the LLM's constraint-modeling capability through a Python solver interface. The LLM receives the problem context and is instructed to formulate the problem using the OR-Tools CP-SAT API~\citep{perron_et_al:LIPIcs.CP.2023.3}.
Rather than explicitly specifying \emph{how} to find a solution, the LLM encodes problem requirements through decision variables, constraints, and optionally an objective function, while the OR-Tools CP-SAT solver automatically performs the search.

\textbf{Declarative constraint modeling (\paradigm{MiniZinc + OR-Tools}).}
This paradigm tests the LLM as a declarative modeler. The LLM receives the problem context and is instructed to generate a MiniZinc model. MiniZinc is a high-level, solver-independent declarative modeling
language for constraint satisfaction and optimization
problems~\citep{nethercote2007minizinc}. 
We adopt the two-step generation pipeline of \citet{song2025llmcp} as a strong MiniZinc baseline, where the LLM is first instructed to generate the model and then a Python formatter that maps the solver's variable assignments into the required format. We use the same CP-SAT solver backend as in the \paradigm{Python + OR-Tools} setting.

% 

% \texttt{solve\_instance(}\allowbreak\texttt{data\_dict)}, but must formulate the
% problem using OR-Tools CP-SAT~\citep{perron_et_al:LIPIcs.CP.2023.3}. The generated
% code constructs variables, constraints, and an objective when
% applicable, while CP-SAT performs the combinatorial search. 

% The model is compiled by the MiniZinc toolchain and solved
% with OR-Tools CP-SAT, matching the \paradigm{Python + OR-Tools}
% backend so that the comparison focuses on the modeling surface rather
% than the solver implementation.

% We adopt the two-step
% MiniZinc generation pipeline of \citet{song2025llmcp}: the LLM emits
% both a MiniZinc model and a paired Python formatter that converts the
% solver's variable assignments into the benchmark's expected output
% dictionary. The model is compiled by the MiniZinc toolchain and solved
% with OR-Tools CP-SAT, matching the \paradigm{Python + OR-Tools}
% back-end so that the comparison focuses on the modeling surface rather
% than the solver implementation. This paradigm tests the LLM as a
% declarative modeler.

For each paradigm, the \textit{baseline} prompt asks for a correct
solver. The \textit{heuristic} prompt additionally suggests classical
efficiency techniques: stronger variable/value ordering, pruning, and
problem decomposition for \paradigm{Python}; tighter bounds, redundant
or implied constraints, symmetry breaking, and explicit search annotations
(e.g.\ \texttt{int\_search} in MiniZinc,
\texttt{AddDecisionStrategy} in CP-SAT) for the two solver-backed
paradigms. The exact prompt texts are given verbatim in
App.~\ref{app:prompts}.

\subsection{Shared Refinement Protocol}
\label{sec:method-refinement}

\begin{figure}[t]
    \centering
    \resizebox{0.96\linewidth}{!}{%
    \begin{tikzpicture}[
        font=\footnotesize,
        node distance=0.62cm,
        box/.style={
            rectangle,
            rounded corners=4pt,
            draw=black!60,
            fill=black!2,
            thick,
            align=center,
            minimum height=0.80cm,
            text width=1.95cm
        },
        smallbox/.style={
            rectangle,
            rounded corners=4pt,
            draw=black!45,
            fill=black!1,
            thick,
            align=center,
            minimum height=0.72cm,
            text width=1.75cm
        },
        emph/.style={
            rectangle,
            rounded corners=4pt,
            draw=blue!60!black,
            fill=blue!6,
            very thick,
            align=center,
            minimum height=0.80cm,
            text width=1.95cm
        },
        artefact/.style={
            draw=orange!75!black,
            fill=orange!12
        },
        arrow/.style={-{Latex[length=2mm]}, thick, draw=black!70},
        feedback/.style={-{Latex[length=2mm]}, thick, dashed, draw=blue!65!black}
    ]
    
    \node[box] (prompt) {\textbf{Prompt}\\problem setup};
    \node[box, right=of prompt] (gen) {\textbf{Generate}\\solver};
    \node[box, right=of gen] (seed) {\textbf{Smoke check}\\(small instance)};
    \node[box, right=of seed] (gate) {\textbf{Scaling check}\\(harder instance)};
    \node[box, artefact, right=of gate] (final) {\textbf{Final}\\artefact};
    \node[emph, right=of final] (eval) {\textbf{Evaluate}\\(all instances)};
    
    \node[smallbox, artefact, below=0.78cm of seed] (first) {\textbf{First-working}\\artefact};
    \node[smallbox, below=0.78cm of gate] (refine) {\textbf{Refine}\\with feedback};
    
    \draw[arrow] (prompt) -- (gen);
    \draw[arrow] (gen) -- (seed);
    \draw[arrow] (seed) -- node[above, inner sep=1pt] {\scriptsize pass} (gate);
    \draw[arrow] (gate) -- node[above, inner sep=1pt] {\scriptsize pass} (final);
    \draw[arrow] (final) -- (eval);
    
    \draw[arrow] (seed) -- node[right, inner sep=1pt] {\scriptsize first pass} (first);
    
    \draw[feedback] (seed.south east) to[out=-35,in=150]
        node[pos=0.4, sloped, above=1pt] {\scriptsize fail} (refine.north west);
    
    \draw[feedback] (gate) -- node[right, inner sep=1pt] {\scriptsize fail} (refine);
    
    \draw[feedback]
        (refine.south) -- ++(0,-0.35)
        -| node[pos=0.25, below, inner sep=1pt] {\scriptsize retry} (gen.south);
    
    \draw[arrow] (first.east) -- ++(0.4,0) -- ++(0,-0.55) -| (eval.south);
    
    \end{tikzpicture}%
    }
    \caption{
    End-to-end workflow applied to the baseline and heuristic settings.
    The \emph{smoke check} verifies that the LLM-generated solver runs and produces a
    valid output on a small instance. %; the first revision to pass is recorded as the \emph{first-working artefact}. 
    The \emph{scaling check} re-runs the artefact on
    a separate mid-difficulty instance under a time budget.
    %; the first revision to pass (or the last refined revision, if none passes within the refinement budget) is retained as the \emph{final artefact}. 
    Failures at either stage are returned as
    feedback for refinement.
    Correctness is verified and runtime recorded during \emph{evaluation}, when both
    first-working and final artefacts are run on \emph{all} instances of the problem.
    %(including the smoke and scaling ones) and their outputs are checked against the    reference solver.
    }
    \label{fig:refinement-workflow}
    \end{figure}
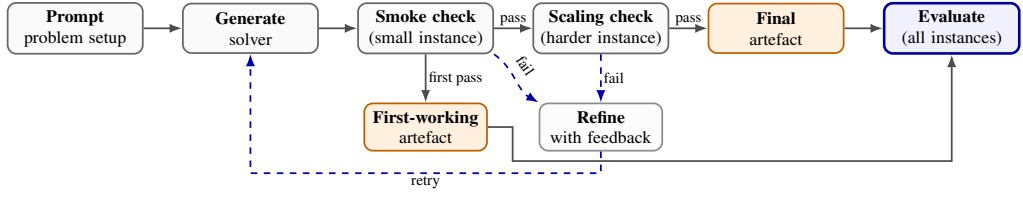
All six (paradigm, prompt) configurations follow the same code-generation pipeline,
summarized in Fig.~\ref{fig:refinement-workflow}. The pipeline applies two validation stages: a
\emph{smoke check} that the artefact runs and emits a schema-valid
output on a small instance, followed by a \emph{scaling check} that it
still completes on a harder instance under a tighter time budget.
Neither stage verifies solution correctness against the reference
solver; SAT/UNSAT checking is deferred to the downstream evaluation
described in Sec.~\ref{sec:exp-metrics}, where both
artefacts produced by the pipeline are run on the full per-problem
instance pool and their outputs are checked against the reference.

The smoke check executes \texttt{solve\_instance(}\allowbreak\texttt{data\_dict)}
for the \paradigm{Python} and \paradigm{Python + OR-Tools} paradigms, and compiles and
solves the model (then runs the formatter) for \paradigm{MiniZinc + OR-Tools};
in both cases the returned dictionary is validated against the
benchmark's expected output schema and value ranges.
The earliest revision to pass the smoke check is recorded as the
\emph{first-working artefact}; the first revision to additionally
pass the scaling check is retained as the \emph{final artefact}, or
the most recent revision if none passes within the refinement budget.
Whenever either check fails, the failure---together with the previous
artefact and the validation feedback---is fed back to the LLM as a
refinement prompt. Each pipeline receives a refinement budget of four
rounds.\footnote{Sampling is greedy throughout ($T=0$, top-$p=1$), and
the model-side reasoning setting is held fixed.}

The protocol is identical for both prompt options: the prompt determines
what the LLM is asked to attempt initially, and refinement iterates that
attempt through the same gates. The scaling check is the workflow's
principal efficiency signal: rejecting artefacts that cannot finish a
harder instance within a tight budget is intended to encourage more
efficient generated solvers.
Empirically, refinement is dominated by basic runnability
recovery---fixing crashes, format violations, and timeouts---not by
speedups on solvers that already run~(Sec.~\ref{sec:trap-refinement}).
The pipeline thus yields, per (LLM, paradigm, prompt, problem) tuple,
a pair of artefacts---the \emph{first-working} and the
\emph{final}---that feed both the experiments (Sec.~\ref{sec:experiments})
and the heuristic-trap analysis~(Sec.~\ref{sec:heuristic-trap}).
Comparing the two artefacts isolates the effect of refinement on
correctness, runtime, and~optimality.

\subsection{Methodological Controls}
\label{sec:method-impl}

Two methodological controls keep the comparison clean.

\textbf{External time control.} All solver runtime budgets are imposed
externally by the evaluation framework, not by the artefact itself.
The prompt explicitly forbids the LLM from setting any internal time
or step limit in the generated code (e.g., assigning to
\texttt{CpSolver().parameters}\allowbreak\texttt{.max\_time\_in\_seconds}) and from
registering callbacks whose purpose is early stopping. Without this
prohibition we observed that LLMs frequently insert their own
per-instance time caps, which would silently truncate
search and confound the paradigm-level comparison.

\textbf{Paradigm enforcement.} The prompt also rules out common ways
the LLM might leave its assigned representational
surface. For \paradigm{Python + OR-Tools}, the LLM must use
\texttt{ortools.sat.python}\allowbreak\texttt{.cp\_model} only, with no
pure-Python fallback or third-party solvers (\texttt{gurobipy}, etc.);
if OR-Tools is unavailable, the artefact must raise a clear
\texttt{RuntimeError}. Without this clause LLMs wrap their
CP-SAT call in a \texttt{try/except} that hands hard instances back to
a hand-rolled Python loop, silently contaminating \paradigm{Python + OR-Tools}
with the native paradigm.
% =====================================================================
% Experiments section for NeurIPS submission.
% Required preamble:
%   \usepackage{booktabs}
%   \usepackage{graphicx}
%   \usepackage{amsmath, amssymb}
%   \usepackage{multirow}
% Figures live in ../neurips_figures/ relative to this file.
% =====================================================================

\section{Experiments}
\label{sec:experiments}

How does a frontier LLM's solving capability change with the
representation it uses for the solver, and with whether the prompt asks
for efficiency? This section answers the first question---and surfaces
the empirical signature of the second, which
Sec.~\ref{sec:heuristic-trap} dissects mechanistically. After
introducing the benchmark and metrics
(Secs.~\ref{sec:exp-benchmark}--\ref{sec:exp-metrics}), we present the
main results across paradigms and the fidelity ranking they
expose (Sec.~\ref{sec:representational}), and close with a five-way
outcome decomposition %by paradigm and prompt
(Sec.~\ref{sec:exp-outcomes}).

\subsection{The CP-SynC-XL Benchmark}
\label{sec:exp-benchmark}

We introduce CP-SynC-XL, an XL-scale extension of the CP-SynC
benchmark~\citep{song2026cpsync}. CP-SynC contains $100$ heterogeneous combinatorial
problems---$58$ Constraint Satisfaction Problems (CSPs) and $42$
Constraint Optimization Problems (COPs)---drawn from CSPLib and the
PyCSP3 problem suite and spanning industrial and academic domains:
sequencing (e.g.\ Car~Sequencing, All-Interval Series), scheduling
(e.g.\ Bus Driver Scheduling, Social Golfers), packing and assignment
(e.g.\ Warehouse Location, Template Design), and combinatorial design
(e.g.\ Golomb Rulers, Quasigroup Existence, Nonogram). Let $\mathcal{P}$
denote the set of problems. Each problem context $p \in \mathcal{P}$
is defined as $p=(P_{\mathrm{NL}}, P_{\mathrm{IN}}, P_{\mathrm{OUT}})$,
where $P_{\mathrm{NL}}$ is the natural-language problem description,
$P_{\mathrm{IN}}$ specifies the typed input parameters, and
$P_{\mathrm{OUT}}$ specifies the required output format; each problem
ships with a reference verifier for CSP feasibility and COP optimality
against a curated reference solver.
The original CP-SynC benchmark evaluates each problem on a single
provided instance solvable by the reference model within $30$ seconds.

To probe solving performance more thoroughly, CP-SynC-XL augments
this with additional parameter instances per problem.
For each problem, we first searched for publicly
available instances from sources including CSPLib, MiniZinc
challenges, and XCSP3 competitions, and converted applicable
instances into the benchmark's input schema $P_{\mathrm{IN}}$,
which yielded external instances for $13$ problems. For the
remaining problems, we wrote problem-specific generators with
curated parameters controlling instance scale, producing instances
at increasing scale (e.g.\ larger $n$ for N-Queens, more customers
and nodes for CVRP). The resulting suite contains $4{,}577$
instances (median $60$ per problem).
The evaluation suite is agnostic to modeling frameworks and
formulations, and checks the solution accuracy of generated solvers
at the output-solution level.
\begin{comment}
A single (LLM, paradigm, prompt, problem) tuple yields one solver
artefact for the declarative paradigm and two artefacts (a
\emph{first-working} and a \emph{final}) for the imperative and native
paradigms (Sec.~\ref{sec:method}); each artefact is then
applied to every parameter instance of its problem under a per-instance
$256$-second wall-clock solver budget. The eighteen
$\langle \text{paradigm},\text{prompt},\text{LLM}\rangle$ triples per
problem produce the per-instance (correct, provided, runtime) records
on which all subsequent figures and tables are computed. The current
section reports the \emph{final} artefacts only; the comparison with
the first-working artefacts, which isolates the marginal value of the
refinement loop, is reported in Sec.~\ref{sec:trap-refinement}.
    
\end{comment}

\subsection{Metrics and Evaluation Protocol}
\label{sec:exp-metrics}

%\paragraph{Solution-level outcomes.}
For every (artefact, instance) pair, the evaluation framework records two binary outcomes: 
(1) \textbf{Provided}, where the artefact returns a schema-conformant output dictionary within the 256\,s budget; and 
(2) \textbf{Correct}, where the provided solution additionally passes the reference verifier (CSP feasibility and COP optimality). 
The ratio \textit{correct}/\textit{provided} is the
\emph{fidelity ratio} of a (paradigm, prompt) pair: the share of
returned solutions that pass the verifier; its complement is the rate
at which the artefact confidently produces a solution that the
verifier rejects. We treat fidelity as a metric of its own in the
analysis below.
We summarise solver speed with cumulative runtime curves \footnote{Curves
are plotted on a base-2 logarithmic time axis from $0.25$\,s to
$T\!=\!256$\,s, with major ticks at the canonical budgets
$\{0.25, 0.5, 1, 2, 4, 8, 16, 32, 64, 128, 256\}$\,s. }
over per-instance solver runtime.
%---\texttt{execution\_seconds}
%for the imperative and native paradigms and \texttt{elapsed\_sec} for
%the declarative paradigm; in both cases this is the wall-clock time of
%the solver subprocess, exclusive of LLM token generation, and is the
%metric the end-user pays. 
For every (LLM, paradigm, prompt) triple and every $t\!\le\!T$ we
report the fraction of instances solved correctly by time $t$. 
\begin{comment}
\paragraph{Reference verifier and inclusion rule.}
The verifier itself runs under a $128$-second per-instance budget. We
treat ``no solution was produced within $256$\,s'' as a legitimate
solver outcome and \emph{keep} those instances; what we cannot keep is
``the verifier could not produce a verdict''. Concretely, an instance
$(\text{problem},q)$ is included in the analyses below if and only if,
in \emph{every} (LLM, paradigm, prompt) triple, its verifier returned a definitive
correct/incorrect verdict (i.e.\ no
\texttt{timeout}/\texttt{worker\_timeout} error and no
``MiniZinc status (opt): UNKNOWN'' on the reference solver).

This single
global rule, applied uniformly across all eighteen triples, removes any
verifier-side-confounded comparisons; under it we retain $2{,}565$
instances drawn from $83$ of the $100$ benchmark problems
(seventeen problems are excluded entirely, either because the reference
solver itself times out or because the verifier returns UNKNOWN under
the $128$-second budget). The included-instance manifest is released
alongside the benchmark.
\end{comment}

\subsection{Results}
\label{sec:representational}

Figure~\ref{fig:km} pairs, for every (LLM, paradigm, prompt) triple,
the \textit{correct} curve (solid) and the \textit{provided}
curve (dotted) on a shared log time axis. The solid curve measures
how often an artefact has returned a verifier-accepted solution by
time $t$; the dotted curve measures how often it has returned any
schema-conformant solution by time $t$. Their vertical gap is
therefore the confident-but-wrong mass, and the ratio between them is
the fidelity summarized in Tab.~\ref{tab:correct-at-256s}. At the
$256$\,s tail, the curves reveal two complementary rankings.

\begin{table}[t]
\centering
\caption{Instance-level outcomes at the $256$\,s budget.
\textbf{C} (correct) = fraction of instances passing the verifier;
\textbf{P} (provided) = fraction returning a schema-conformant
output; \textbf{F} (fidelity) = $\textbf{C}/\textbf{P}$, the share
of returned solutions that pass the verifier.
Bold = best per LLM on each metric.}
\label{tab:correct-at-256s}
\footnotesize
\setlength{\tabcolsep}{3pt}
\begin{tabular}{ll rrr rrr rrr}
\toprule
& & \multicolumn{3}{c}{\textsc{DeepSeek-V3.2}}
  & \multicolumn{3}{c}{\textsc{GPT-5.3-Codex}}
  & \multicolumn{3}{c}{\textsc{Gemini 3.1 Pro}} \\
\cmidrule(lr){3-5}\cmidrule(lr){6-8}\cmidrule(lr){9-11}
Paradigm & Prompt
  & \multicolumn{1}{c}{C} & \multicolumn{1}{c}{P} & \multicolumn{1}{c}{F}
  & \multicolumn{1}{c}{C} & \multicolumn{1}{c}{P} & \multicolumn{1}{c}{F}
  & \multicolumn{1}{c}{C} & \multicolumn{1}{c}{P} & \multicolumn{1}{c}{F} \\
\midrule
\paradigm{Python}              & baseline  & $0.291$          & $0.478$          & $60.9\%$          & $0.634$          & $0.781$          & $81.2\%$          & $0.606$          & $0.708$          & $85.6\%$          \\
\paradigm{Python}              & heuristic & $0.295$          & $\mathbf{0.568}$ & $51.9\%$          & $0.630$          & $0.781$          & $80.7\%$          & $0.640$          & $0.761$          & $84.1\%$          \\
\paradigm{Python + OR-Tools}   & baseline  & $\mathbf{0.439}$ & $0.523$          & $83.9\%$          & $0.702$          & $0.786$          & $89.3\%$          & $0.716$          & $\mathbf{0.816}$ & $87.8\%$          \\
\paradigm{Python + OR-Tools}   & heuristic & $0.365$          & $0.538$          & $67.8\%$          & $\mathbf{0.711}$ & $\mathbf{0.802}$ & $88.7\%$          & $\mathbf{0.722}$ & $0.814$          & $88.7\%$          \\
\paradigm{MiniZinc + OR-Tools} & baseline  & $0.300$          & $0.338$          & $\mathbf{88.8\%}$ & $0.522$          & $0.565$          & $\mathbf{92.4\%}$ & $0.544$          & $0.588$          & $\mathbf{92.5\%}$ \\
\paradigm{MiniZinc + OR-Tools} & heuristic & $0.269$          & $0.323$          & $83.3\%$          & $0.541$          & $0.597$          & $90.6\%$          & $0.508$          & $0.570$          & $89.1\%$          \\
\bottomrule
\end{tabular}
\end{table}

\textbf{Correctness ranks \paradigm{Python + OR-Tools} $>$ \paradigm{Python} $\gtrsim$
\paradigm{MiniZinc + OR-Tools} on every LLM.} On every LLM, the
\paradigm{Python + OR-Tools} correct curve is higher than both other
paradigms over essentially the whole time axis, and the best
(paradigm, prompt) pair \emph{per LLM} at the $256$\,s tail uses
\paradigm{Python + OR-Tools}. The advantage is not that the LLM
performs more combinatorial reasoning in Python; rather, delegating
search to CP-SAT moves the main burden from designing a complete
algorithm to encoding the variables, constraints, and objective
faithfully. \paradigm{MiniZinc + OR-Tools}, by contrast, falls below
\paradigm{Python + OR-Tools} on every LLM and below \paradigm{Python} on \textsc{DeepSeek-V3.2},
despite carrying the largest theoretical speed-up surface (global
constraints, search annotations). Because both solver-backed
paradigms use OR-Tools CP-SAT, this gap is consistent with a
surface-fluency bottleneck rather than a weaker back-end.\footnote{The
construct frequencies of Sec.~\ref{sec:trap-frequency} make this
interpretation more concrete.}

\textbf{Fidelity ranks \paradigm{MiniZinc + OR-Tools} $>$ \paradigm{Python + OR-Tools} $>$
\paradigm{Python} on every LLM.} The \textbf{F} column of
Tab.~\ref{tab:correct-at-256s}---the share of returned solutions that
pass the verifier, equivalently the ratio of the solid curve to the
dotted curve in Fig.~\ref{fig:km}---inverts the correctness ranking.
\paradigm{MiniZinc + OR-Tools} maintains $\geq\!83\%$ fidelity on every (LLM, prompt) pair;
\paradigm{Python + OR-Tools} is intermediate (down to $65.0\%$ on
\textsc{DeepSeek-V3.2}'s heuristic prompt); \paradigm{Python} is lowest
(bottoming at $51.9\%$ on the same pair). This ranking follows from
what it means for each paradigm to provide an answer. In
\paradigm{Python}, a returned solution is simply the output of a
hand-written procedure. In \paradigm{Python + OR-Tools}, it is a
CP-SAT-feasible assignment to the model that the LLM encoded. In
\paradigm{MiniZinc + OR-Tools}, it is a CP-SAT-feasible assignment to
a model that first had to clear MiniZinc's type and well-formedness
checks. Each step narrows the class of errors that can be hidden by a
schema-conformant return value. \paradigm{MiniZinc + OR-Tools} therefore has the
highest conditional fidelity when it answers, despite its lower
overall correct rate. A natural future direction is therefore
to train or adapt models with stronger fluency in declarative modeling
languages such as MiniZinc, rather than treating the current gap as an
intrinsic limitation of declarative solver construction.

\textbf{The heuristic prompt yields marginal, inconsistent gains.} The
heuristic prompt asks the LLM to improve efficiency without changing
the problem semantics; empirically, the effect is much weaker and less
stable than this goal suggests. On the stronger LLMs (\textsc{GPT-5.3-Codex} and
\textsc{Gemini 3.1 Pro}) every paradigm-by-metric change stays within
$\pm\!3.6$ percentage points (pp) of the baseline---a mix of small
positives and small negatives, with no paradigm consistently helped
across both LLMs. On the
weakest LLM (\textsc{DeepSeek-V3.2}) the prompt degrades performance:
$-7.4$\,pp of correctness and $-16.1$\,pp of fidelity on
\paradigm{Python + OR-Tools}, plus $-9.0$\,pp of fidelity on
\paradigm{Python}. Both halves of this pattern---marginal,
inconsistently-signed effects on the stronger LLMs and sizeable,
uniformly negative losses on the weakest---are the empirical
signature of the \emph{heuristic trap};
Sec.~\ref{sec:heuristic-trap} traces them to specific code-level
patterns the LLM injects that plausibly speed up the model on only a fraction of instances
while introducing mis-encoding bugs.

\subsection{Outcome Decomposition by Paradigm and Prompt}
\label{sec:exp-outcomes}

To identify where the provided--correct gap comes from, we decompose every
(artefact, instance) outcome (Fig.~\ref{fig:outcomes}) into five
mutually exclusive bins: \emph{correct}, where the verifier accepts
the solution; \emph{suboptimal}, a feasible COP solution worse than
the optimum; \emph{UNSAT}, where the LLM-generated solver returns a
schema-conformant solution but pinning that solution into the
reference verifier yields an \texttt{UNSAT} verdict on an instance the
reference solver finds feasible; \emph{invalid}, a returned-and-rejected case, typically a CSP
solution violating a mis-encoded constraint; and \emph{no solution}, no schema-conformant output
within the $256$\,s budget (timeout, crash, or schema violation).
Figure~\ref{fig:outcomes} shows three recurring patterns:

\begin{figure}[t]
\centering
\includegraphics[width=\textwidth]{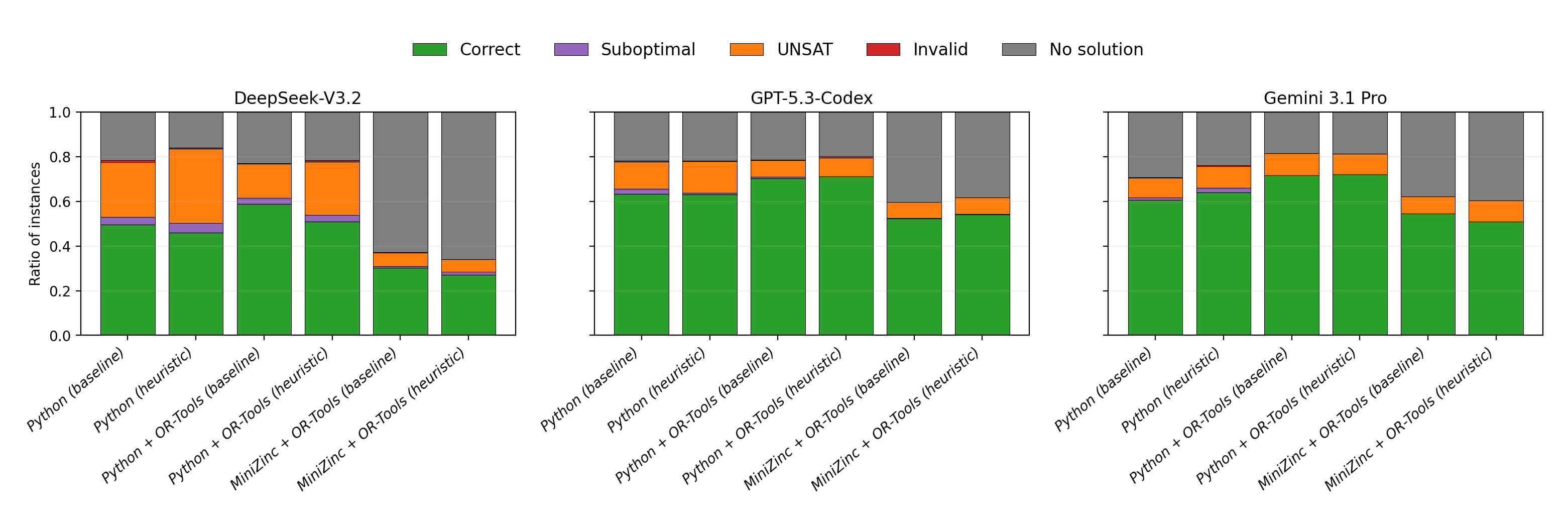}
\caption{Five-way outcome decomposition by paradigm and prompt, one
panel per LLM. The bins partition the shared instance pool:
\emph{correct} (verifier accepts); \emph{suboptimal} (feasible COP
solution above the optimum); \emph{UNSAT} (the returned solution,
when pinned into the reference verifier, yields an \texttt{UNSAT}
verdict on a feasible instance);
\emph{invalid} (returned-and-rejected, typically a violated
constraint); \emph{no solution} (no schema-conformant output within
$256$\,s).}
\label{fig:outcomes}
\end{figure}

\textbf{No-solution is the largest non-correct slice on every panel.}
Inspection of the underlying logs shows these are overwhelmingly
$256$\,s timeouts rather than crashes or malformed outputs---the
budget ceiling compounds with whatever encoding errors each paradigm
contributes.

\textbf{Suboptimal is a \paradigm{Python} failure mode.}
The suboptimal bin is essentially absent from the solver-backed
paradigms---where CP-SAT either proves the optimum or times
out---but visible on every \paradigm{Python} panel, where
hand-rolled COP procedures return feasible-but-non-optimal answers
at the budget---an approximation attempt rather than guaranteed
optimization.

\textbf{\emph{UNSAT} is the dominant wrong verdict on the solver-backed
paradigms.} Once the larger no-solution share of
\paradigm{MiniZinc + OR-Tools} is netted out, both
\paradigm{Python + OR-Tools} and \paradigm{MiniZinc + OR-Tools}
concentrate their wrong-mass on \emph{UNSAT} rather than
\emph{invalid}: the generated model admits returned assignments that
do not satisfy the reference formulation. A manual audit of the top
twelve problems by aggregated \emph{UNSAT} mass on the
\paradigm{Python + OR-Tools} \textit{baseline} setting ($750$ of
$837$ instances, $89.6\%$, across all three LLMs) shows that every
audited case is a natural-language-to-formal-language
mistranslation. We call these baseline mistranslation mechanisms
\textbf{M1--M4}:
%; Sec.~\ref{sec:heuristic-trap} uses a separate
%\textbf{A--F} notation for heuristic-prompt regression modes. The
%baseline mechanisms are:
\textbf{M1}~associative recall (the LLM names the problem before it
reads it, e.g.\ a Hamiltonian template imposed on the
Travelling-Purchaser problem);
\textbf{M2}~long-spec attention drift (one or two textual
constraints are silently dropped);
\textbf{M3}~output-schema ambiguity between LLM and verifier (the
LLM's decision-variable wiring does not round-trip through the
reference); and
\textbf{M4}~sentinel-value misreading (overloaded zero/negative
markers in the input are encoded as ``no edge'' or
``unreachable''). M2 ($38\%$ of the audited \emph{UNSAT} mass) and M3
($36\%$) jointly account for $\approx\!75\%$ of the audited mass,
with M1 ($16\%$) and M4 ($10\%$) covering the rest: the dominant
baseline failure is not solver-side reasoning but a fluency-shaped
reading of the natural-language specification. App.~\ref{sec:baseline-modes} presents the audit per problem
together with the offending code extracts.

% =====================================================================
% Analysis section for NeurIPS submission.
% Required preamble:
%   \usepackage{booktabs}
%   \usepackage{listings}
%   \usepackage{xcolor}
%   \usepackage{amsmath, amssymb}
%   \usepackage{multirow}
% Recommended listing style for inline source code:
%   \lstdefinestyle{pyinline}{
%     basicstyle=\ttfamily\footnotesize,
%     keywordstyle=\color{black!70}\bfseries,
%     commentstyle=\color{gray}\itshape,
%     breaklines=true,
%     showstringspaces=false,
%   }
% =====================================================================

\section{Anatomy of the Heuristic Trap}
\label{sec:heuristic-trap}

To understand \emph{why} the \textit{heuristic} prompt is unsafe on
some (LLM, paradigm) pairs, we performed a paired code-level analysis
across all three paradigms and three LLMs. This section uses
\textit{baseline} ($B$) and \textit{heuristic} ($H$) to denote the two
prompt conditions. Correctness changes are reported as
$\Delta\text{acc}=\text{acc}_{H}-\text{acc}_{B}$ in percentage points
unless a table explicitly reports fractions; speed changes are
reported as $t_B/t_H$, so values above $1$ mean the heuristic-prompt
solver is faster. Two findings organise the section. First,
net-positive correctness gains and visible speed-ups concentrate on a
narrow set of (LLM, paradigm) pairs. Second, the remaining regressions
trace back to six paradigm-specific failure modes whose net effect
depends on the LLM.

\subsection{Aggregate Heuristic Effects}
\label{sec:trap-aggregate}

For \textsc{GPT-5.3-Codex}, the macro-averaged correctness change
$\Delta\text{acc}$ (\textit{heuristic} minus \textit{baseline},
percentage points) is mildly positive on every paradigm
($+1.2$, $+1.7$, $+2.2$\,pp on \paradigm{Python},
\paradigm{Python + OR-Tools}, \paradigm{MiniZinc + OR-Tools}); for
\textsc{Gemini 3.1 Pro} it oscillates around zero
($+2.2$, $-0.6$, $-1.1$\,pp); for \textsc{DeepSeek-V3.2} it regresses
on every paradigm, by as much as $-6.6$\,pp on
\paradigm{Python + OR-Tools}. Thus, the same heuristic instruction can
be mildly beneficial for one model and harmful for another; aggregating
over LLMs would obscure this model-specific effect.
The runtime picture is similarly selective. On \paradigm{Python} and
\paradigm{Python + OR-Tools} the median per-problem speed-up
$t_B/t_H$ is $\approx 1.0\times$ for every LLM,
and $40$--$63\%$ of common-correct instances are in fact slower
under the \textit{heuristic} prompt. \paradigm{MiniZinc + OR-Tools}
is the only paradigm where heuristics operate at the model level
so CP-SAT can exploit them
deterministically; everywhere else the LLM's heuristic suggestions
either re-express constraints or substitute its own search procedure,
and CP-SAT's propagation already does the work the LLM is trying to
add.

\subsection{Six Failure Modes Behind the Regressions}
\label{sec:trap-modes}

A code-level audit of heuristic-prompt solvers identifies
\textbf{six failure modes (A--F)} that cover the dominant recurring
sources of correctness regressions in our audit. These modes are
separate from the baseline mistranslation mechanisms M1--M4 in
Sec.~\ref{sec:exp-outcomes}: M1--M4 explain why a baseline solver can
return a wrong answer, whereas A--F explain what the
\textit{heuristic} prompt newly changes relative to the paired
baseline. Each mode is marked by a conservative source-level detector
and then interpreted through paired outcome changes.
\textbf{A}~unverified ``tight'' bounds
(\paradigm{Python + OR-Tools}; $13.5\%$ incidence);
\textbf{B}~Big-$M$ / MTZ rewrites that weaken propagation
(\paradigm{Python + OR-Tools}; $4.0\%$);
\textbf{C}~hand-computed seeding
(\paradigm{Python + OR-Tools}; $19.5\%$);
\textbf{D}~silent loss of completeness (\paradigm{Python}; $16.2\%$);
\textbf{E}~COP$\to$SAT collapse against an unproven optimum
(\paradigm{MiniZinc + OR-Tools}; $0.7\%$); and
\textbf{F}~overcomplicated redundant-machinery blow-ups
(\paradigm{MiniZinc + OR-Tools}; $29.1\%$).
Per-LLM incidence and within-mode outcome split appear in
Tab.~\ref{tab:failure-mode-outcome}. Full mechanism descriptions,
detector definitions, and source-level code extracts are discussed in
App.~\ref{app:trap-modes-details}.

\begin{table}[t]
\centering
\caption{Failure-mode incidence and within-mode outcome split, per
LLM (relevant-paradigm population $N\!\approx\!99$ per LLM, slightly
smaller for \paradigm{MiniZinc + OR-Tools} where some problems do
not compile). 
\textbf{Inc.}\ is the fraction of that population whose
heuristic-prompt solver matches the mode's detector;
$\overline{\Delta\text{acc}}$ is the within-mode mean correctness
change reported as a fraction (e.g.\ $+0.028=+2.8$\,pp);
\textbf{\%\,reg./\%\,imp.}\ partition within-mode problems by
$\Delta\text{acc}$ ($\leq\!-5$\,pp / $\geq\!+5$\,pp);
\textbf{Med.\ sp.}\ is the median $t_B/t_H$ on common-correct
instances, and \textbf{\%f / \%s} is the percentage of those instances
with $t_B/t_H>1$ / $<1$. Bold rows: the (mode, LLM) pair produces a
net regression \emph{or} a $>\!25\%$ within-mode regression rate.
Modes A--D and F use behavioural source-level detectors that are
LLM-agnostic; Mode~E uses a stricter ``base optimises but heuristic
satisfies'' diff and so reflects an explicit COP$\to$SAT collapse
rather than an LLM-rewritten objective bound.
}
\label{tab:failure-mode-outcome}
\footnotesize
\setlength{\tabcolsep}{3pt}
\resizebox{\linewidth}{!}{
\begin{tabular}{*{10}{c}}
\toprule
\textbf{Mode} & \textbf{Paradigm} & \textbf{LLM} &
$n_{\text{mode}}$ & Inc.\ &
$\overline{\Delta\text{acc}}$ & \%\,reg. & \%\,imp.\ & Med.\ sp.\ &
\%f / \%s \\
\midrule
\multirow{3}{*}{\textbf{A}} & \multirow{3}{*}{\paradigm{Python + OR-Tools}}   & \textsc{GPT-5.3-Codex}          & $11$           & $11.1$           & $+0.028$           & $\phantom{0}9.1$ & $18.2$ & $0.96$ & $\phantom{0}0.4 / \phantom{0}3.5$ \\
                            &                                                & \textsc{Gemini 3.1 Pro}         & $10$           & $10.1$           & $+0.022$           & $\phantom{0}0.0$ & $10.0$ & $0.99$ & $\phantom{0}4.2 / \phantom{0}1.8$ \\
                            &                                                & \textbf{\textsc{DeepSeek-V3.2}} & $\mathbf{19}$  & $\mathbf{19.2}$  & $\mathbf{-0.067}$  & $\mathbf{21.1}$  & $26.3$ & $0.95$ & $\phantom{0}0.0 / \phantom{0}4.7$ \\
\midrule
\multirow{3}{*}{\textbf{B}} & \multirow{3}{*}{\paradigm{Python + OR-Tools}}   & \textsc{GPT-5.3-Codex}          & $\phantom{0}2$ & $\phantom{0}2.0$ & $+0.150$           & $\mathbf{50.0}$  & $50.0$ & $0.88$ & $20.5 / \mathbf{59.1}$ \\
                            &                                                & \textbf{\textsc{Gemini 3.1 Pro}}& $\phantom{0}3$ & $\phantom{0}3.0$ & $-0.078$           & $\mathbf{33.3}$  & $\phantom{0}0.0$ & $1.00$ & $\phantom{0}2.3 / \phantom{0}2.3$ \\
                            &                                                & \textbf{\textsc{DeepSeek-V3.2}}& $\phantom{0}7$ & $\phantom{0}7.1$ & $+0.029$           & $\mathbf{28.6}$  & $28.6$ & $1.01$ & $37.3 / \phantom{0}3.9$ \\
\midrule
\multirow{3}{*}{\textbf{C}} & \multirow{3}{*}{\paradigm{Python + OR-Tools}}   & \textsc{GPT-5.3-Codex}          & $26$           & $26.3$           & $+0.009$           & $15.4$           & $23.1$         & $1.00$ & $\phantom{0}3.8 / \phantom{0}5.2$ \\
                            &                                                & \textsc{Gemini 3.1 Pro}         & $\phantom{0}9$ & $\phantom{0}9.1$ & $+0.018$           & $\phantom{0}0.0$ & $11.1$         & $1.02$ & $\phantom{0}6.8 / \phantom{0}0.0$ \\
                            &                                                & \textsc{DeepSeek-V3.2}          & $23$           & $23.2$           & $+0.061$           & $\phantom{0}8.7$ & $17.4$         & $0.96$ & $\phantom{0}2.9 / \phantom{0}1.3$ \\
\midrule
\multirow{3}{*}{\textbf{D}} & \multirow{3}{*}{\paradigm{Python}}              & \textsc{GPT-5.3-Codex}          & $14$           & $14.1$           & $+0.018$           & $\phantom{0}7.1$ & $14.3$ & $0.95$ & $\phantom{0}4.2 / 17.7$ \\
                            &                                                & \textsc{Gemini 3.1 Pro}         & $11$           & $11.1$           & $+0.055$           & $\phantom{0}9.1$ & $18.2$ & $1.00$ & $\phantom{0}0.0 / \phantom{0}1.7$ \\
                            &                                                & \textbf{\textsc{DeepSeek-V3.2}}& $\mathbf{23}$  & $\mathbf{23.2}$  & $\mathbf{-0.037}$  & $\mathbf{21.7}$  & $17.4$ & $0.97$ & $\phantom{0}0.7 / \phantom{0}0.7$ \\
\midrule
\multirow{3}{*}{\textbf{E}} & \multirow{3}{*}{\paradigm{MiniZinc + OR-Tools}} & \textbf{\textsc{GPT-5.3-Codex}}& $\phantom{0}2$ & $\phantom{0}2.0$ & $\mathbf{-0.334}$  & $\mathbf{50.0}$  & $\phantom{0}0.0$ & $0.88$ & $41.2 / \phantom{0}5.9$ \\
                            &                                                & \textsc{Gemini 3.1 Pro}         & $\phantom{0}0$ & $\phantom{0}0.0$ & ---                & ---              & ---            & ---    & --- \\
                            &                                                & \textsc{DeepSeek-V3.2}          & $\phantom{0}0$ & $\phantom{0}0.0$ & ---                & ---              & ---            & ---    & --- \\
\midrule
\multirow{3}{*}{\textbf{F}} & \multirow{3}{*}{\paradigm{MiniZinc + OR-Tools}} & \textsc{GPT-5.3-Codex}          & $\mathbf{39}$  & $\mathbf{39.8}$  & $+0.013$           & $17.9$           & $23.1$ & $1.17$ & $44.0 / 13.5$ \\
                            &                                                & \textsc{Gemini 3.1 Pro}         & $11$           & $11.3$           & $+0.009$           & $18.2$           & $18.2$ & $1.05$ & $25.9 / 15.7$ \\
                            &                                                & \textbf{\textsc{DeepSeek-V3.2}}& $\mathbf{30}$  & $\mathbf{37.5}$  & $+0.002$           & $\mathbf{30.0}$  & $30.0$ & $0.90$ & $30.8 / 26.4$ \\
\bottomrule
\end{tabular}
}
\end{table}

Two patterns cut across the modes; the per-LLM
$\overline{\Delta\text{acc}}$ and within-mode regression rates cited
below are taken from Tab.~\ref{tab:failure-mode-outcome}.
\textbf{Mode~A's net effect depends on bound quality, not bound presence.}
Mode~A is net-positive for \textsc{GPT-5.3-Codex} and
\textsc{Gemini 3.1 Pro} but net-negative for \textsc{DeepSeek-V3.2}:
\textsc{GPT-5.3-Codex}'s ``bound'' comments are typically correct
theorems in disguise, whereas \textsc{DeepSeek-V3.2}'s are more often
wishful invariants. The resulting swing in $\overline{\Delta\text{acc}}$
($-6.7$\,pp on \textsc{DeepSeek-V3.2} vs.\ $+2.8$\,pp on
\textsc{GPT-5.3-Codex}) reflects the \emph{verifiability} of what
each LLM annotates as a bound, not the mode itself.
\textbf{The most-frequent mode is bimodal; the rare modes are
high-risk.} Mode~F drives both \paradigm{MiniZinc + OR-Tools}'s
genuine speed-ups and its heaviest regressions: the redundant
auxiliary structure sometimes carries propagation that CP-SAT cannot
derive on its own, and sometimes overwhelms flattening or tightens
domains too aggressively---an outcome we cannot predict before the
run. Modes~B and~E are rare in absolute incidence but cripple
correctness when triggered ($-33.4$\,pp mean for~E on
\textsc{GPT-5.3-Codex}; $33\%$ within-mode regression rate for~B).

\section{Conclusion}
\label{sec:conclusion}

We studied LLM solver synthesis along two axes: representation and
search optimization. Across three frontier LLMs, \textsc{GPT-5.3-Codex}, \textsc{Gemini 3.1 Pro}, and \textsc{DeepSeek-V3.2},
\paradigm{Python + OR-Tools} attains the highest correctness despite
sharing CP-SAT backend with \paradigm{MiniZinc + OR-Tools}, while native
\paradigm{Python} is the most likely to return schema-valid but
verifier-rejected answers. This pattern is consistent with the
modeling surface itself being a central bottleneck, not only the
underlying solver.
On the optimization axis, heuristic prompting yields modest
\paradigm{MiniZinc + OR-Tools} speed-ups but has a bimodal effect:
some generated artefacts improve, while others become slower or less
correct. A paired code audit traces these regressions to six
paradigm-specific failure modes, with the risk concentrated when the
LLM replaces solver-native machinery with unverified bounds, hard
fixes, or algorithmic substitutions. The practical implication is to
use the LLM primarily at the formalisation interface---variables,
constraints, objectives, and carefully scoped solver-native
annotations---and to subject LLM-authored search optimizations to
independent validation.

\bibliographystyle{plainnat} % 这是最常用的 NeurIPS 引用格式
\bibliography{neurips_2026}  % your_filename 是你的 .bib 文件名（不带扩展名）

%%%%%%%%%%%%%%%%%%%%%%%%%%%%%%%%%%%%%%%%%%%%%%%%%%%%%%%%%%%%

\appendix

\section{Limitations}
\label{app:limitations}

Three controls bound the external validity of our findings.
(i)~We evaluate three frontier LLMs from a late-$2025$ snapshot;
future or smaller models may exhibit different paradigm preferences,
and absolute fidelity numbers should be read as a snapshot rather
than an upper bound. (ii)~We fix OR-Tools as the back-end across
both solver-backed paradigms, so the representational-divergence
finding should be read as evidence for this CP-SAT-backed
API/declarative pairing rather than as a claim about all MIP-,
SAT-, or CP-modeling pipelines. (iii)~We operate in
the single-shot synthesis regime with $\leq\!4$ refinement rounds
and a $256$\,s solver budget per instance; the heuristic-trap claims
therefore concern bounded-refinement deployment settings rather than
the population-and-selection regimes used by FunSearch-style
algorithm-synthesis loops. We do not study finetuning, tool-augmented
agents, or multi-instance amortised solving, all of which lie
outside the scope of this paper.

\section{Prompt details and refinement templates}
\label{app:prompts}

We list, verbatim, the prompts and refinement templates used by the
three paradigms of Sec.~\ref{sec:method-paradigms}. Each preset ships
exactly the text below; placeholders \texttt{\{err\_msg\}},
\texttt{\{previous\_code\}}, and \texttt{\{refinement\_feedback\_message\}}
are substituted from the evaluation framework's state at refinement time. Sampling is
greedy ($T\!=\!0$, top-$p\!=\!1$); for models that expose a reasoning
budget we fix the budget at the medium tier across all generators.
LLM calls are made through the OpenRouter chat-completions endpoint
\texttt{https://openrouter.ai/api/v1/chat/completions}. The three
model identifiers used in the experiments are
\texttt{openai/gpt-5.3-codex},
\texttt{google/gemini-3.1-pro-preview}, and
\texttt{deepseek/deepseek-v3.2}. Each request uses a system message
and user/refinement messages, forwards \texttt{max\_tokens=128000},
\texttt{temperature=0}, \texttt{top\_p=1}, and, when supported by the
provider, \texttt{reasoning.effort=medium}; the request timeout is
$300$\,s and transient HTTP/network failures are retried with
exponential backoff. The run configuration persisted with every
generated artefact records the exact model id, prompt/backend preset,
refinement budget, sampling parameters, package versions, and the
repository commit.

\subsection{Python-Based Pipelines (\paradigm{Python} and \paradigm{Python + OR-Tools})}
\label{app:prompts-py}

\paragraph{System prompt.}

\begin{promptbox}
You are an expert constraint programming and algorithm design assistant.
You will be given:
- A problem name and full description (and optionally input/output specs).
- How the inputs are structured: parameters arrive as a Python dict
  `data_dict`. The prompt may describe keys by type/schema (not fixed
  concrete values) because the same code must solve many instances of
  this problem.

Your task is to write Python code that **defines** a function:

    def solve_instance(data_dict) -> dict:
        ...
        return hypothesis_solution

Requirements:
- Do NOT read or write files, do NOT use the network.
- Read all problem data from `data_dict` at runtime (do not assume one
  fixed instance).
- Do not print anything or execute the solver; just define the function.
- At the bottom of your code, do NOT call `solve_instance`; we will
  call it.

Output ONLY valid Python code. Do not include explanations or comments
outside code.
\end{promptbox}

\paragraph{User prompt template.}
The user prompt concatenates the natural-language problem description,
an inferred schema for the parameter dictionary, the list of required
output keys, and the two toggle blocks below. The full schema and
required-output blocks are synthesized from the released benchmark and
are omitted here for brevity.

\paragraph{Algorithmic guidance --- \textit{baseline} preset.}

\begin{promptbox}
Algorithmic guidance: produce a correct solver.
\end{promptbox}

\paragraph{Algorithmic guidance --- \textit{heuristic} preset.}

\begin{promptbox}
Algorithmic guidance:
- Use heuristics whenever possible to improve search efficiency.
- Avoid naive exhaustive DFS/backtracking as the primary approach.
- Design for scalability: strong variable/value ordering, pruning,
  symmetry breaking, feasibility prechecks, decomposition, and
  restart/randomized multi-start when useful.
- For optimization problems, propagate tight objective bounds and use
  incumbent-based pruning (valid dominance / bound reasoning), not
  wall-clock or arbitrary time budgets.
- Include concise internal helper structure that makes pruning logic
  explicit.
\end{promptbox}

\paragraph{Backend constraints --- \paradigm{Python + OR-Tools}.}

\begin{promptbox}
Backend preference:
- Use OR-Tools CP-SAT (`ortools.sat.python.cp_model`) for modeling.
- Do NOT configure CP-SAT wall-clock or step limits that stop search
  early. Forbidden examples (non-exhaustive): assigning to
  `CpSolver().parameters.max_time_in_seconds`,
  `solver.parameters.max_time_in_seconds`, or similar time/step caps;
  time-based callbacks whose purpose is early stopping. Let the outer
  evaluation framework control runtime.
- STRICT requirement: You must solve using OR-Tools CP-SAT only.
- Do NOT use gurobipy and do NOT use a pure-Python fallback solver.
- If OR-Tools is unavailable, raise a clear RuntimeError immediately.
- Keep the public entrypoint unchanged: `solve_instance(data_dict) -> dict`.
\end{promptbox}

\paragraph{Backend constraints --- native paradigm.}

\begin{promptbox}
Backend preference:
- Use Python standard library only.
- Do not rely on third-party optimization libraries.
- Do NOT set any explicit time limit inside generated code
  (e.g., no solver-style TimeLimit and no custom early-stop
  wall-clock cutoff logic).
\end{promptbox}

\paragraph{Refinement template.}
Whenever a generated solver fails either the seed-instance check or the
$30$\,s gate check, the LLM is re-prompted with the template below; the
algorithmic-guidance and backend blocks are appended unchanged.

\begin{promptbox}
The following Python solver code needs refinement based on
runtime/validation feedback. Fix the code and output ONLY the corrected
Python code (no explanation). Define exactly one function:
`solve_instance(data_dict) -> dict`.

Feedback:
```
{refinement_feedback_message}
```

Previous code:
```python
{previous_code}
```

Keep the same function signature and continue following these strategy
requirements:
{algorithmic guidance for the active preset}

{backend block for the active paradigm}

Provide the corrected code in a single markdown code block. Do not call
`solve_instance` at the bottom; we will call it.
\end{promptbox}

\subsection{Declarative Pipeline}
\label{app:prompts-mzn}

\paragraph{System prompt --- \textit{baseline}-MiniZinc.}

\begin{promptbox}
You are an expert in modeling constraint programming tasks with
MiniZinc. Read the problem description in markdown, then model this
problem and provide MiniZinc code for your model.
\end{promptbox}

\paragraph{System prompt --- \textit{heuristic}-MiniZinc.}
The baseline system prompt is augmented with the following sentence:

\begin{promptbox}
Prioritize model efficiency as well as correctness. Prefer formulations
that solve efficiently, using tighter bounds, redundant constraints,
symmetry breaking, implied constraints, search guidance, or similar
techniques when helpful.
\end{promptbox}

\paragraph{User prompt for stage~1.}
The user prompt concatenates the problem description, the data-file
schema, and a format directive declaring the required output variables.
The latter follows the ``two-stage with format constraints'' template of
\citet{song2025llmcp}; we leave its phrasing unchanged for
comparability with that pipeline. The full rendered prompts, including
the problem-specific schema and required-output block, are saved in the
released run logs together with the generated MiniZinc model and output
formatter.

\paragraph{Stage-3 user prompt --- output formatter.}

\begin{promptbox}
The model has been successfully executed, and the decision variable
values have been stored. Briefly explain the task and write a Python
function named `transform_n_display` that processes these solutions,
stores them in a dictionary in the specified format, and then returns
this dictionary. The function should take two dictionaries as
parameters: (1) `data_dict`, which contains all the input parameter
values, and (2) `decision_var_dict`, which contains exactly the
decision variable values as used in the MiniZinc model you generated.
[...] Ensure your Python script handles these transformations
accurately and returns the dictionary with variables in the following
format: {required output spec}.
\end{promptbox}

\paragraph{Refinement template --- syntax / runtime error.}

\begin{promptbox}
I have executed the code you just generated and it caused an error
message from the IDE:
```
{err_msg}
```

Please review the error message and the code, explain why this error
occurred in one sentence, then fix the error and provide the correct
code in a markdown code block.
\end{promptbox}

\paragraph{Refinement template --- efficiency-gate failure
(\textit{heuristic}-MiniZinc only).}

\begin{promptbox}
I ran the model on another benchmark instance with a 30-second time
limit and it did not meet the required solving target:
```
{err_msg}
```

Please improve the model efficiency and provide the full corrected code
in a markdown code block. Keep the model syntax-correct. You can add
redundant constraints, symmetry breaking, tighter bounds, implied
constraints, search guidance, or any other tricks that narrow the
search space.
\end{promptbox}

\paragraph{Refinement template --- evaluation feedback.}

\begin{promptbox}
The code you just provided has been executed, and the generated
solution was checked by the CPMP evaluation script. It returned the
following result or error:
```
{err_msg}
```

Please review the evaluation feedback, explain the cause in one
sentence, and then fix the code and provide the correct code in a
markdown code block.
\end{promptbox}

\paragraph{Computing infrastructure.}
All experiments were executed on a Linux compute pool of dual-socket
x86\_64 servers running openSUSE Leap~16.0 with kernel~6.12. Each
node is equipped with two Intel Xeon Gold processors from the Xeon
Scalable family (16~cores per socket, 2.1--2.3\,GHz base /
3.7--3.9\,GHz turbo, 22\,MiB L3 cache per socket), exposing
64~hardware threads ($2\!\times\!16$ physical cores with 2-way
simultaneous multithreading) distributed over two NUMA domains.
Each node is provisioned with at least~$345\,$GiB of ECC DDR4 main
memory, comfortably above the resident-set requirement of any
constraint-programming subprocess we spawn. The CPU frequency
governor was pinned to \texttt{performance} on every node to
eliminate dynamic frequency scaling as a confounder for runtime
measurements.

The constraint-programming back-end is MiniZinc~2.9.6 driving
Google OR-Tools CP\nobreakdash-SAT~9.15, invoked from Python~3.11.15
through the official \texttt{minizinc-python} bindings. Each
generated model is replayed against every parameter instance under a
strict $256$-second per-instance solver budget, matching the
generation-time budget used by the LLM code\nobreakdash-refinement
loop; instance \emph{verification} against the reference checker
operates under a tighter $128$-second budget. To bound resource
contention, every node executes at most~$100$ concurrent solver
subprocesses, with the driver enforcing both a per-task estimated
memory reservation and a global guard that refuses to launch a new
worker if free system memory would fall below $10\%$ of total RAM.

We pin every (LLM, generation\nobreakdash-strategy) experiment to a
single physical node, so per-instance runtimes reported
within an experiment are mutually consistent and free of
cross-machine drift. All scripts, raw per-instance solver logs,
generated solution pickles, and evaluation transcripts are released
alongside the paper to permit byte-level reproduction.

\subsection{Experimental details}
\label{asec:exp_details}
\paragraph{Instance filtering.}
Due to time and compute constraints, we impose a $256$ s runtime limit when validating each candidate instance with the reference solver. Candidate instances whose reference validation does not complete within this limit, including those marked as \texttt{UNKNOWN} or ending in timeout, are excluded from the reported comparisons. This filtering leaves $2{,}565$ instances in the final evaluation set reported in the figures and tables.

\subsection{Backend-Specific Construct Families and How Often They Appear}
\label{sec:trap-frequency}

We extracted, from every LLM-generated solver in the heuristic preset,
the set of constructs the heuristic prompt could plausibly elicit.
Table~\ref{tab:patterns-per-backend} reports the percentage of problems
($N\!\approx\!99$ per pair) whose heuristic-prompt solver contains each
construct, for every $(\textit{LLM}, \text{backend})$ pair. The
construct families are very different across backends---\emph{search
hooks} dominate \paradigm{Python + OR-Tools}, \emph{search annotations and globals} dominate
\paradigm{MiniZinc + OR-Tools}, and \emph{algorithmic substitutions} dominate \paradigm{Python}---which
is itself a finding: when the same prompt is delivered, the same LLM
chooses different syntactic surfaces depending on the language it is
instructed to emit.

\begin{table}[t]
\centering
\caption{Frequency (\%) with which each heuristic construct appears in
the LLM's heuristic-prompt solver, across $99$ problems per (LLM, backend) pair. Bold
values flag the dominant construct(s) per backend per LLM. Three
sub-tables, one per backend.}
\label{tab:patterns-per-backend}
\footnotesize
\setlength{\tabcolsep}{4pt}
\begin{tabular}{lrrr}
\toprule
\multicolumn{4}{c}{\textbf{Pure Python} (Native Algorithmic Search)} \\
\midrule
\textbf{Construct} & \textsc{GPT-5.3-Codex} & \textsc{Gemini 3.1 Pro} & \textsc{DeepSeek-V3.2} \\
\midrule
DFS / backtracking                          & $\mathbf{45.5}$ & $\mathbf{51.5}$ & $\mathbf{44.4}$ \\
Memoisation (\texttt{lru\_cache}/\texttt{@cache}) & $\phantom{0}3.0$ & $\phantom{0}4.0$ & $\phantom{0}2.0$ \\
DP table                                    & $\phantom{0}7.1$ & $\phantom{0}3.0$ & $\phantom{0}7.1$ \\
Branch \& bound                             & $\phantom{0}2.0$ & $\phantom{0}8.1$ & $\phantom{0}9.1$ \\
Beam / A$^{\star}$                          & $\phantom{0}3.0$ & $\phantom{0}5.1$ & $\phantom{0}2.0$ \\
Local search (SA / 2-opt / hill-climb)      & $\phantom{0}6.1$ & $\phantom{0}0.0$ & $\phantom{0}4.0$ \\
Multi-start / random restart                & $\phantom{0}5.1$ & $\phantom{0}6.1$ & $\mathbf{18.2}$ \\
Randomised choice (\texttt{random.*})       & $\phantom{0}0.0$ & $13.1$           & $\mathbf{37.4}$ \\
Iteration-cap truncation                    & $\phantom{0}7.1$ & $\phantom{0}2.0$ & $\mathbf{12.1}$ \\
Symmetry breaking                           & $13.1$           & $14.1$           & $20.2$ \\
Dominance / bound (in comments)             & $\phantom{0}8.1$ & $17.2$           & $\mathbf{30.3}$ \\
\midrule
\multicolumn{4}{c}{\textbf{OR-Tools CP-SAT} (\paradigm{Python + OR-Tools})} \\
\midrule
\texttt{AddDecisionStrategy} (var/val ord.)        & $\mathbf{71.7}$ & $\phantom{0}5.1$ & $20.2$ \\
\texttt{AddHint} (warm start)                       & $22.2$           & $\phantom{0}5.1$ & $21.2$ \\
\texttt{num\_search\_workers}                       & $20.2$           & $11.1$           & $\mathbf{64.6}$ \\
\texttt{random\_seed}                               & $\phantom{0}3.0$ & $\phantom{0}0.0$ & $17.2$ \\
Custom objective bound (\texttt{Add(} obj $\le k$)) & $\mathbf{58.6}$  & $\mathbf{40.4}$  & $\mathbf{52.5}$ \\
\texttt{AddCircuit}                                 & $\phantom{0}6.1$ & $\phantom{0}9.1$ & $\phantom{0}4.0$ \\
\texttt{AddMultiplicationEquality}                  & $\phantom{0}7.1$ & $\phantom{0}6.1$ & $\phantom{0}7.1$ \\
\texttt{OnlyEnforceIf} (reified)                    & $19.2$           & $27.3$           & $26.3$ \\
Big-$M$ / MTZ encoding                              & $\phantom{0}0.0$ & $\phantom{0}1.0$ & $\phantom{0}7.1$ \\
Symmetry-breaking constraints                       & $32.3$           & $\mathbf{43.4}$  & $\mathbf{63.6}$ \\
Variable fixing / hard equalities                   & $20.2$           & $12.1$           & $10.1$ \\
Pre-step Python heuristic fed to CP-SAT             & $\phantom{0}2.0$ & $\phantom{0}0.0$ & $\phantom{0}0.0$ \\
\midrule
\multicolumn{4}{c}{\textbf{MiniZinc} (Declarative Modeling)} \\
\midrule
Custom search annotation                            & $\mathbf{89.8}$ & $22.7$ & $\mathbf{48.7}$ \\
\quad \texttt{first\_fail}                          & $\mathbf{70.4}$ & $14.4$ & $25.0$ \\
\quad \texttt{dom\_w\_deg}                          & $20.4$ & $\phantom{0}3.1$ & $\phantom{0}3.7$ \\
\quad \texttt{input\_order}                         & $27.6$ & $\phantom{0}9.3$ & $18.8$ \\
\quad \texttt{smallest}/\texttt{largest}            & $\phantom{0}8.2$ & $12.4$ & $17.5$ \\
\texttt{all\_different}                             & $33.7$ & $32.0$ & $\mathbf{42.5}$ \\
\texttt{global\_cardinality}                        & $\phantom{0}7.1$ & $\phantom{0}6.2$ & $\phantom{0}0.0$ \\
\texttt{circuit}/\texttt{subcircuit}                & $\phantom{0}5.1$ & $\phantom{0}7.2$ & $\phantom{0}1.3$ \\
\texttt{cumulative}/\texttt{disjunctive}            & $\phantom{0}4.1$ & $\phantom{0}3.1$ & $\phantom{0}2.5$ \\
\texttt{lex\_lesseq}/\texttt{value\_precede}        & $21.4$ & $11.3$ & $16.3$ \\
\texttt{table}/\texttt{regular}                     & $\phantom{0}4.1$ & $\phantom{0}3.1$ & $\phantom{0}1.3$ \\
Redundant / implied constraint (in comments)        & $\mathbf{76.5}$ & $\mathbf{46.4}$ & $\mathbf{62.5}$ \\
Symmetry-breaking constraint (lex / value-prec.)    & $\mathbf{60.2}$ & $37.1$           & $\mathbf{58.8}$ \\
Prefix-/running-sum auxiliary                       & $\phantom{0}7.1$ & $\phantom{0}2.1$ & $\phantom{0}0.0$ \\
Tighter explicit variable bounds                    & $\mathbf{75.5}$ & $\mathbf{75.3}$ & $\mathbf{61.3}$ \\
\bottomrule
\end{tabular}
\end{table}

Two patterns in Tab.~\ref{tab:patterns-per-backend} are worth
flagging. First, \textsc{GPT-5.3-Codex} reaches for native CP-SAT
search hooks under the heuristic prompt ($71.7\%$
\texttt{AddDecisionStrategy} on \paradigm{Python + OR-Tools},
$70.4\%$ \texttt{first\_fail} on \paradigm{MiniZinc + OR-Tools}),
which the next subsection shows are largely safe.
Second, \textsc{DeepSeek-V3.2} reaches for higher-risk devices at much
higher rates---randomised choice ($37.4\%$ vs.\ $0.0\%$),
multi-restart ($18.2\%$ vs.\ $5.1\%$), and iteration caps ($12.1\%$
vs.\ $7.1\%$) on \paradigm{Python}; $64.6\%$
\texttt{num\_search\_workers} and $63.6\%$ symmetry-breaking
constraints on \paradigm{Python + OR-Tools} (vs.\
\textsc{GPT-5.3-Codex}'s $20.2\%$ and $32.3\%$). These same
construct families are over-represented among its within-pattern
regressions.

\subsection{The Trap's Signature: Per-Pattern Outcome Breakdown}
\label{sec:trap-bypattern}

A construct's \emph{frequency} alone tells us nothing about its
\emph{effect}. To isolate effect we condition on the construct:
for every problem in which the heuristic-prompt solver contains
construct $C$, we record the paired correctness change
$\Delta\text{acc}=\text{acc}_{H}-\text{acc}_{B}$ relative to the
baseline-prompt solver and, on common-correct instances, the
per-instance speed ratio $t_B/t_H$. We aggregate within
$(\textit{LLM},\,\text{backend},\,C)$. Table~\ref{tab:trap-by-pattern}
reports the constructs whose effect was most informative; we focus on
constructs with $\geq\!5$ supporting problems and on the
correctness-regression rate (within-pattern).

\begin{table}[H]
\centering
\caption{Per-pattern outcome breakdown. ``Coverage'' is the fraction of
$N\!\approx\!99$ problems whose heuristic-prompt solver contains the
construct (same as Table~\ref{tab:patterns-per-backend}). ``\%\,regress'',
``\%\,improve'' are the within-pattern fractions of problems with
$\Delta\text{acc}\!\leq\!-5$\,pp resp.\ $\geq\!+5$\,pp. ``Med.\ sp.'' is
the median per-problem $t_B/t_H$ on
common-correct instances; ``\%\,faster'', ``\%\,slower'' are the within-pattern
fractions of common-correct \emph{instances} (not problems) with
ratio~$\geq\!1.2$ resp.\ $\leq\!0.8$. Bold rows: regression-rate
$\geq\!25\%$ \emph{or} mean $\Delta\text{acc}\leq\!-5$\,pp.}
\label{tab:trap-by-pattern}
\footnotesize
\setlength{\tabcolsep}{3.5pt}
\resizebox{\textwidth}{!}{%
\begin{tabular}{lllrrrrrrr}
\toprule
\textbf{LLM} & \textbf{Backend} & \textbf{Construct} & Cov.\ &
$\overline{\Delta\text{acc}}$ & \%\,reg. & \%\,impr.\ & Med.\ sp.\ &
\%\,faster & \%\,slower \\
\midrule
\multicolumn{10}{l}{\textit{(a) Mostly safe constructs (within-pattern regression \(<\!15\%\))}} \\
\midrule
\textsc{GPT-5.3-Codex}      & OR-Tools & \texttt{AddDecisionStrategy}                  & $71.7$ & $+0.013$ & $\phantom{0}8.5$ & $14.1$ & $1.00$ & $\phantom{0}2.9$ & $\phantom{0}6.9$ \\
\textsc{GPT-5.3-Codex}      & OR-Tools & \texttt{OnlyEnforceIf}                         & $19.2$ & $+0.109$ & $\phantom{0}5.3$ & $31.6$ & $1.00$ & $\phantom{0}3.0$ & $\phantom{0}1.0$ \\
\textsc{GPT-5.3-Codex}      & OR-Tools & symmetry break (lex / sort)                    & $32.3$ & $+0.063$ & $\phantom{0}3.1$ & $18.8$ & $1.00$ & $\phantom{0}0.2$ & $\phantom{0}2.4$ \\
\textsc{GPT-5.3-Codex}      & MiniZinc & \texttt{circuit}/\texttt{subcircuit}           & $\phantom{0}5.1$ & $+0.017$ & $\phantom{0}0.0$ & $20.0$ & $\mathbf{5.54}$ & $\mathbf{100.0}$ & $\phantom{0}0.0$ \\
\textsc{GPT-5.3-Codex}      & MiniZinc & \texttt{all\_different}                        & $33.7$ & $+0.078$ & $\phantom{0}9.1$ & $27.3$ & $1.40$ & $53.9$ & $10.0$ \\
\textsc{GPT-5.3-Codex}      & MiniZinc & \texttt{first\_fail}                           & $70.4$ & $+0.011$ & $13.0$ & $15.9$ & $1.21$ & $55.5$ & $11.0$ \\
\textsc{Gemini 3.1 Pro} & MiniZinc & \texttt{circuit}/\texttt{subcircuit}           & $\phantom{0}7.2$ & $+0.024$ & $14.3$ & $28.6$ & $\mathbf{1.46}$ & $\phantom{0}5.1$ & $83.5$\textsuperscript{\dag} \\
\textsc{Gemini 3.1 Pro} & Python   & DFS / backtracking                              & $51.5$ & $+0.037$ & $13.7$ & $27.5$ & $1.00$ & $\phantom{0}6.8$ & $\phantom{0}2.9$ \\
\textsc{GPT-5.3-Codex}      & MiniZinc & redundant/implied constraint                   & $76.5$ & $+0.030$ & $16.0$ & $20.0$ & $1.18$ & $52.3$ & $13.9$ \\
\midrule
\multicolumn{10}{l}{\textit{(b) High-risk ``trap'' constructs (within-pattern regression \(\geq\!25\%\) or $\overline{\Delta\text{acc}}\leq\!-5$\,pp)}} \\
\midrule
\textbf{\textsc{GPT-5.3-Codex}} & MiniZinc & \texttt{smallest}/\texttt{largest} val.\ sel.    & $\phantom{0}8.2$ & $\mathbf{-0.225}$ & $\mathbf{62.5}$ & $\phantom{0}0.0$ & $1.42$ & $48.4$ & $17.2$ \\
\textbf{\textsc{GPT-5.3-Codex}} & MiniZinc & \texttt{global\_cumulative}                      & $\phantom{0}4.1$ & $\mathbf{-0.188}$ & $25.0$           & $\phantom{0}0.0$ & $1.23$ & $44.0$ & $\phantom{0}6.0$ \\
\textbf{\textsc{GPT-5.3-Codex}} & MiniZinc & \texttt{global\_lex} / \texttt{value\_precede}   & $21.4$           & $+0.027$         & $\mathbf{33.3}$  & $14.3$ & $1.08$ & $42.1$ & $10.7$ \\
\textbf{\textsc{GPT-5.3-Codex}} & OR-Tools & \texttt{random\_seed}                            & $\phantom{0}3.0$ & $\mathbf{-0.315}$ & $\mathbf{33.3}$  & $\phantom{0}0.0$ & $0.97$ & $\phantom{0}0.0$ & $\phantom{0}0.0$ \\
\textbf{\textsc{GPT-5.3-Codex}} & Python   & dominance/bound (claimed in comments)           & $\phantom{0}8.1$ & $\mathbf{-0.076}$ & $12.5$           & $25.0$ & $0.98$ & $\phantom{0}0.0$ & $12.3$ \\
\textbf{\textsc{Gemini 3.1 Pro}} & MiniZinc & \texttt{global\_lex} / \texttt{value\_precede}      & $11.3$ & $-0.006$         & $\mathbf{36.4}$ & $27.3$ & $1.08$ & $23.4$ & $15.6$ \\
\textbf{\textsc{Gemini 3.1 Pro}} & MiniZinc & \texttt{dom\_w\_deg}                                  & $\phantom{0}3.1$ & $-0.063$         & $\mathbf{33.3}$ & $\phantom{0}0.0$ & $1.14$ & $45.5$ & $27.3$ \\
\textbf{\textsc{Gemini 3.1 Pro}} & OR-Tools & \texttt{circuit\_constraint}                          & $\phantom{0}9.1$ & $-0.050$         & $22.2$ & $11.1$ & $1.00$ & $\phantom{0}1.9$ & $\phantom{0}0.6$ \\
\textbf{\textsc{DeepSeek-V3.2}}  & OR-Tools & \texttt{AddDecisionStrategy}                          & $20.2$ & $\mathbf{-0.108}$ & $\mathbf{35.0}$ & $15.0$ & $0.99$ & $\phantom{0}5.6$ & $\phantom{0}6.3$ \\
\textbf{\textsc{DeepSeek-V3.2}}  & OR-Tools & \texttt{OnlyEnforceIf}                                & $26.3$ & $\mathbf{-0.141}$ & $\mathbf{30.8}$ & $\phantom{0}3.8$ & $0.97$ & $\phantom{0}6.4$ & $\phantom{0}5.8$ \\
\textbf{\textsc{DeepSeek-V3.2}}  & OR-Tools & \texttt{num\_search\_workers}                         & $\mathbf{64.6}$ & $\mathbf{-0.070}$ & $\mathbf{26.6}$ & $18.8$ & $0.99$ & $\phantom{0}8.4$ & $\phantom{0}3.5$ \\
\textbf{\textsc{DeepSeek-V3.2}}  & OR-Tools & symmetry-breaking constraint                          & $\mathbf{63.6}$ & $\mathbf{-0.090}$ & $\mathbf{27.0}$ & $12.7$ & $0.99$ & $\phantom{0}8.2$ & $\phantom{0}4.0$ \\
\textbf{\textsc{DeepSeek-V3.2}}  & OR-Tools & variable fixing / hard equality                       & $10.1$ & $\mathbf{-0.138}$ & $\mathbf{30.0}$ & $\phantom{0}0.0$ & $1.01$ & $12.0$ & $\phantom{0}1.7$ \\
\textbf{\textsc{DeepSeek-V3.2}}  & MiniZinc & \texttt{first\_fail}                                  & $25.0$ & $\mathbf{-0.118}$ & $25.0$           & $25.0$ & $0.88$ & $27.7$ & $37.1$ \\
\textbf{\textsc{DeepSeek-V3.2}}  & MiniZinc & \texttt{global\_lex} / \texttt{value\_precede}        & $16.3$ & $\mathbf{-0.063}$ & $\mathbf{46.2}$ & $38.5$ & $0.86$ & $44.7$ & $25.5$ \\
\textbf{\textsc{DeepSeek-V3.2}}  & MiniZinc & redundant/implied constraint                          & $\mathbf{62.5}$ & $\mathbf{-0.053}$ & $\mathbf{32.0}$ & $26.0$ & $0.90$ & $24.9$ & $31.3$ \\
\textbf{\textsc{DeepSeek-V3.2}}  & Python   & iteration-cap truncation                              & $12.1$           & $\mathbf{-0.073}$ & $\mathbf{41.7}$ & $16.7$ & $0.95$ & $\phantom{0}0.0$ & $\phantom{0}1.6$ \\
\textbf{\textsc{DeepSeek-V3.2}}  & Python   & dominance/bound (claimed in comments)                 & $\mathbf{30.3}$  & $+0.024$          & $\mathbf{26.7}$ & $16.7$ & $0.99$ & $\phantom{0}5.5$ & $\phantom{0}0.4$ \\
\textbf{\textsc{DeepSeek-V3.2}}  & Python   & DFS / backtracking                                    & $\mathbf{44.4}$  & $-0.002$          & $\mathbf{22.7}$ & $18.2$ & $0.97$ & $13.8$ & $\phantom{0}3.4$ \\
\bottomrule
\end{tabular}
}
{\footnotesize\flushleft \textsuperscript{\dag} For a single problem the
heuristic-prompt circuit pruning is mathematically valid for small
boards but unsound for the largest instances; the median speed-up is
high because the easy half is much faster, while the hard half times out.}
\end{table}

Two patterns emerge from the per-pattern conditioning.

\textbf{Construct families have intrinsic risk profiles}, but those
profiles \emph{interact} with model capability. \texttt{first\_fail} is a
canonical safe heuristic in CP, and indeed for \textsc{GPT-5.3-Codex} it regresses on only
$13.0\%$ of \paradigm{MiniZinc + OR-Tools} problems (mean $\Delta\text{acc}\!=\!+0.011$); for
\textsc{DeepSeek-V3.2} the same construct regresses on $25.0\%$ of problems with mean
$\Delta\text{acc}\!=\!-0.118$ and a median \emph{slow}-down of $0.88\times$.
The construct is identical---the difference is whether the LLM
applies it on a variable scope that admits propagation or on a
mis-modelled scope that does not. \texttt{AddDecisionStrategy},
\texttt{OnlyEnforceIf}, and symmetry-breaking constraints exhibit the
same pattern in \paradigm{Python + OR-Tools}: $5\%\!-\!10\%$ regression for \textsc{GPT-5.3-Codex} and
$25\%\!-\!35\%$ for \textsc{DeepSeek-V3.2} (Table~\ref{tab:trap-by-pattern}, panel b).

\textbf{A subset of constructs is intrinsically dangerous.} Three
constructs regress at high rates regardless of LLM:
\begin{itemize}\itemsep0pt
    \item \emph{Aggressive value selectors} on \paradigm{MiniZinc + OR-Tools}
    (\texttt{smallest}/\texttt{largest}, \texttt{dom\_w\_deg}): \textsc{GPT-5.3-Codex}
    regresses on $62.5\%$ of \texttt{smallest}/\texttt{largest} problems
    by $-22.5$\,pp on average---the worst within-pattern regression in our
    benchmark. The selector imposes a value-ordering decision that, when
    the variable's domain ordering does not align with feasibility,
    forecloses on the satisfying region.
    \item \emph{Symmetry / lex-precede chains beyond the trivial scope}:
    on \paradigm{MiniZinc + OR-Tools}, \texttt{value\_precede\_chain} regresses on $33\%\!-\!46\%$
    of problems for all three LLMs---the constraint is correct only when
    the values are interchangeable, a property the LLM rarely verifies.
    \item \emph{Iteration caps in pure Python} ($+\!\textsc{DeepSeek-V3.2}$
    only): \texttt{max\_iter} cuts off the search before the search has
    explored the feasible region; $41.7\%$ of problems regress.
\end{itemize}

These three families are the clearest examples of constructs that violate the
problem's underlying \emph{semantics} (rather than its search ordering or
propagation) and on which the LLM cannot self-verify.

\section{Refinement Efficacy Statistics}
\label{app:refinement-stats}
\label{sec:trap-refinement}

The pipeline retains both the \emph{first-working} version of each
generated solver---the earliest refinement attempt to pass a
seed-instance runtime and format check---and the
\emph{final-refined} version after up to four refinement rounds aimed
at clearing a $30$\,s execution gate.
Tab.~\ref{tab:first-vs-final} compares the two artefacts on the
shared instance pool. The comparison yields three conclusions.
\textbf{(R1)~Refinement does not measurably accelerate solvers}: the
median speed-up is essentially $1.0\times$ in every (paradigm,
prompt) pair, and on the \texttt{heuristic\_ortools} setting $51.9\%$
of overlapping instances are slower after refinement.
\textbf{(R2)~Refinement is an error-recovery loop}: the modest
correctness gains ($+1.4$ to $+3.5$\,pp on three of four pairs) are
driven almost entirely by ``final-only'' instances---cases that
crashed or violated the format check on the first round---rather than
by speed-ups on already-correct instances.
\textbf{(R3)~Refinement under the \textit{heuristic} prompt is
partially counterproductive}: the \texttt{heuristic\_py} setting
shows $\Delta\text{acc}\!=\!-0.2$\,pp ($99$ instances become
first-only and only $89$ become final-only), and
\texttt{heuristic\_ortools} shows the worst \%\,slow rate of any
setting ($51.9\%$). When the prompt encourages additional
efficiency-oriented machinery, refinement helps least, and on
\paradigm{Python} it slightly hurts: each refinement round tends to
preserve the already-injected heuristic rather than remove it. The same
trend manifests in \paradigm{MiniZinc + OR-Tools} through Modes~E
and~F (App.~\ref{app:trap-modes-details}), where the LLM keeps
strengthening an already-broken model in successive rounds.

\begin{table}[h]
\centering
\caption{Refinement effect: first-working artefact versus the final artefact, on a shared instance pool ($N{\approx}4{,}500$). ``Speed-up (first/final)'' is the per-instance ratio over instances where both artefacts are correct; values $>\!1$ mean the final solver is faster.}
\label{tab:first-vs-final}
\small
\setlength{\tabcolsep}{4pt}
\begin{tabular}{lrrrrrrr}
\toprule
\textbf{Preset} & $N$ & $\text{acc}_{\text{first}}$ & $\text{acc}_{\text{final}}$ &
$\Delta\text{acc}$ & $\widetilde{\text{sp}}$ & $\overline{\text{sp}}$ & \%\,slow \\
\midrule
baseline\_py        & $4517$ & $0.403$ & $0.418$ & $+1.5$\,pp           & $1.03\times$ & $1.05\times$ & $37.8\%$ \\
\textbf{heuristic\_py}      & $4517$ & $0.431$ & $0.429$ & $\mathbf{-0.2}$\,\textbf{pp}     & $1.01\times$ & $1.33\times$ & $\mathbf{45.2\%}$ \\
baseline\_ortools   & $4457$ & $0.435$ & $0.471$ & $+3.5$\,pp           & $1.01\times$ & $1.02\times$ & $42.9\%$ \\
heuristic\_ortools  & $4517$ & $0.454$ & $0.468$ & $+1.4$\,pp           & $0.99\times$ & $0.98\times$ & $\mathbf{51.9\%}$ \\
\bottomrule
\end{tabular}
\end{table}

\section{Correctness at Canonical Time Budgets}
\label{app:budget-bars}

\begin{figure}[H]
\centering
\includegraphics[width=\textwidth]{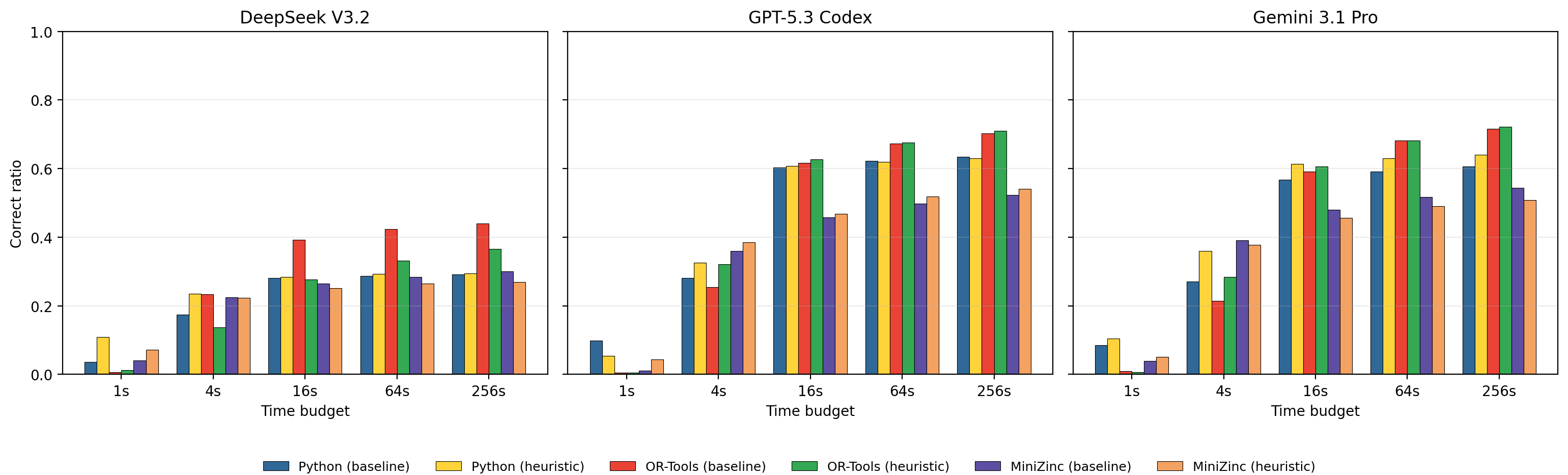}
\caption{Correct ratio at canonical time budgets
$\{1, 4, 16, 64, 256\}$\,s, for each LLM (panels) and each
(paradigm, prompt) setting (six bars per budget). The pattern at
every budget mirrors the cumulative runtime curves of Fig.~\ref{fig:km}:
\paradigm{Python + OR-Tools} dominates,
\paradigm{MiniZinc + OR-Tools} lags, and the
\textit{heuristic}--\textit{baseline} relation flips across LLMs.}
\label{fig:budget-bars}
\end{figure}

Fig.~\ref{fig:budget-bars} provides a discrete-budget complement to
the cumulative runtime curves of Fig.~\ref{fig:km}: for each LLM and each canonical
budget $\{1, 4, 16, 64, 256\}$\,s, it reports the correct ratio
across the six (paradigm, prompt) settings. The pattern at every
budget mirrors the cumulative curves: \paradigm{Python + OR-Tools} dominates,
\paradigm{MiniZinc + OR-Tools} lags, and the
\textit{heuristic}--\textit{baseline} relation flips across LLMs.
The figure is most useful for cross-referencing the construct- and
mode-level discussion of Sec.~\ref{sec:heuristic-trap}: most per-mode
regressions appear in the $4$\,--\,$64$\,s window, where simple
solvers have already terminated and harder ones have not yet timed
out, and the mid-budget bars surface those regressions more legibly
than the tail of the cumulative curve.

\section{Per-Problem Cumulative Runtime Curves}
\label{app:per-problem-km}

\begin{figure}[H]
\centering
\includegraphics[width=\textwidth]{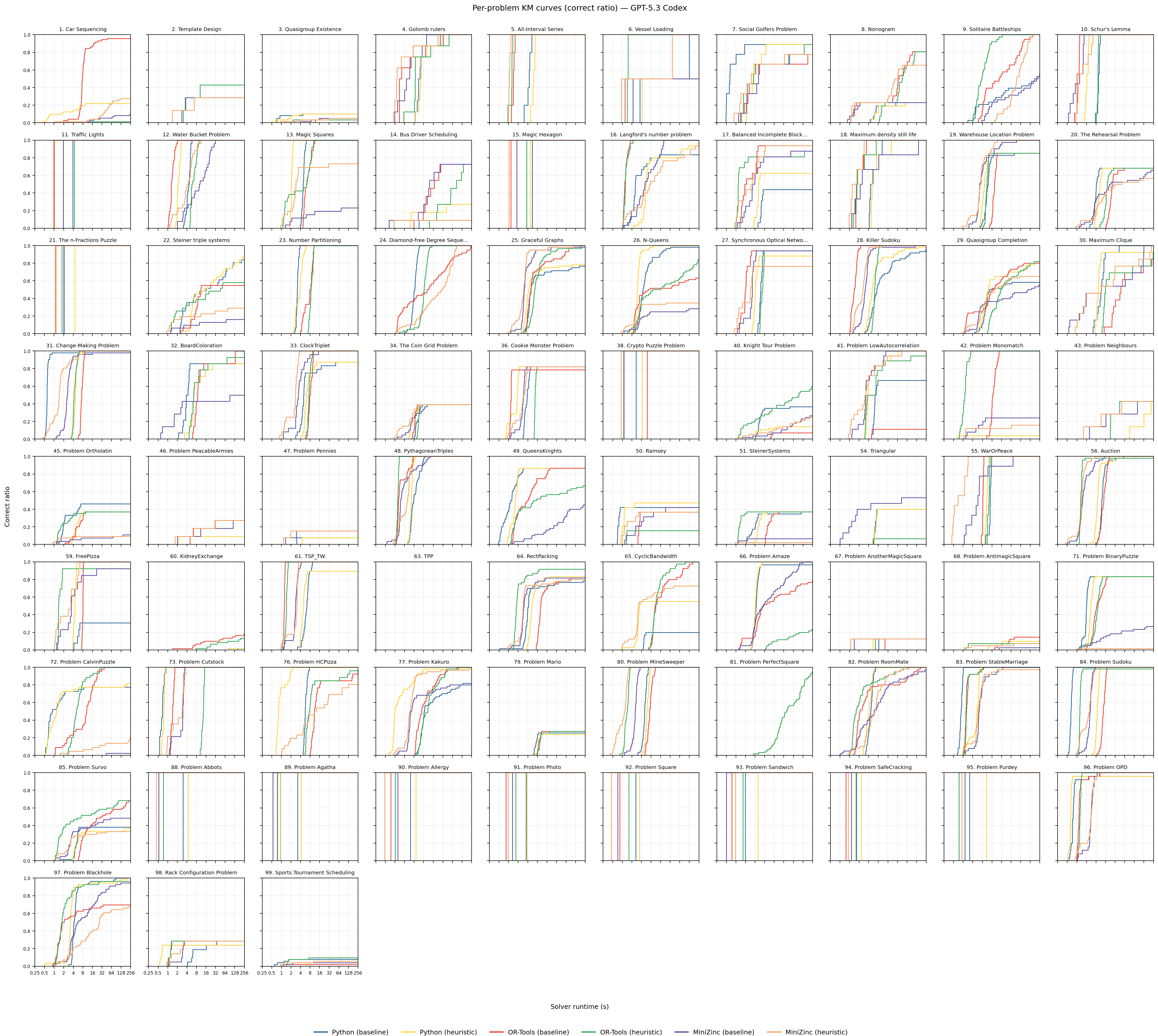}
\caption{Per-problem cumulative runtime curves (correct ratio) for
\textsc{GPT-5.3-Codex}, with one subplot per problem and six curves
per subplot ($3$ paradigms $\times$ $2$ prompts).}
\label{fig:per-problem-codex}
\end{figure}

\begin{figure}[H]
\centering
\includegraphics[width=\textwidth]{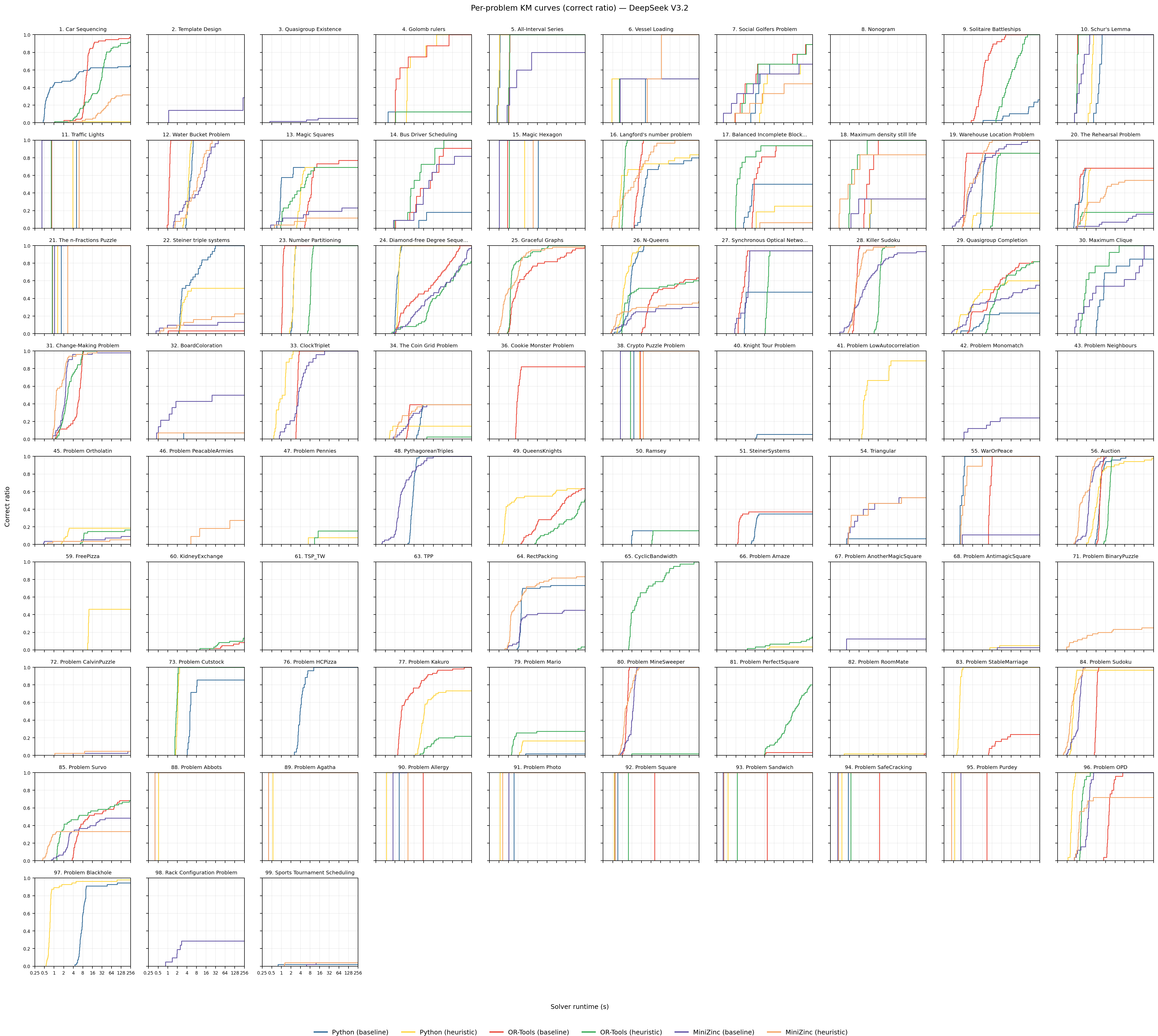}
\caption{Per-problem cumulative runtime curves (correct ratio) for
\textsc{DeepSeek-V3.2}.}
\label{fig:per-problem-deepseek}
\end{figure}

\begin{figure}[H]
\centering
\includegraphics[width=\textwidth]{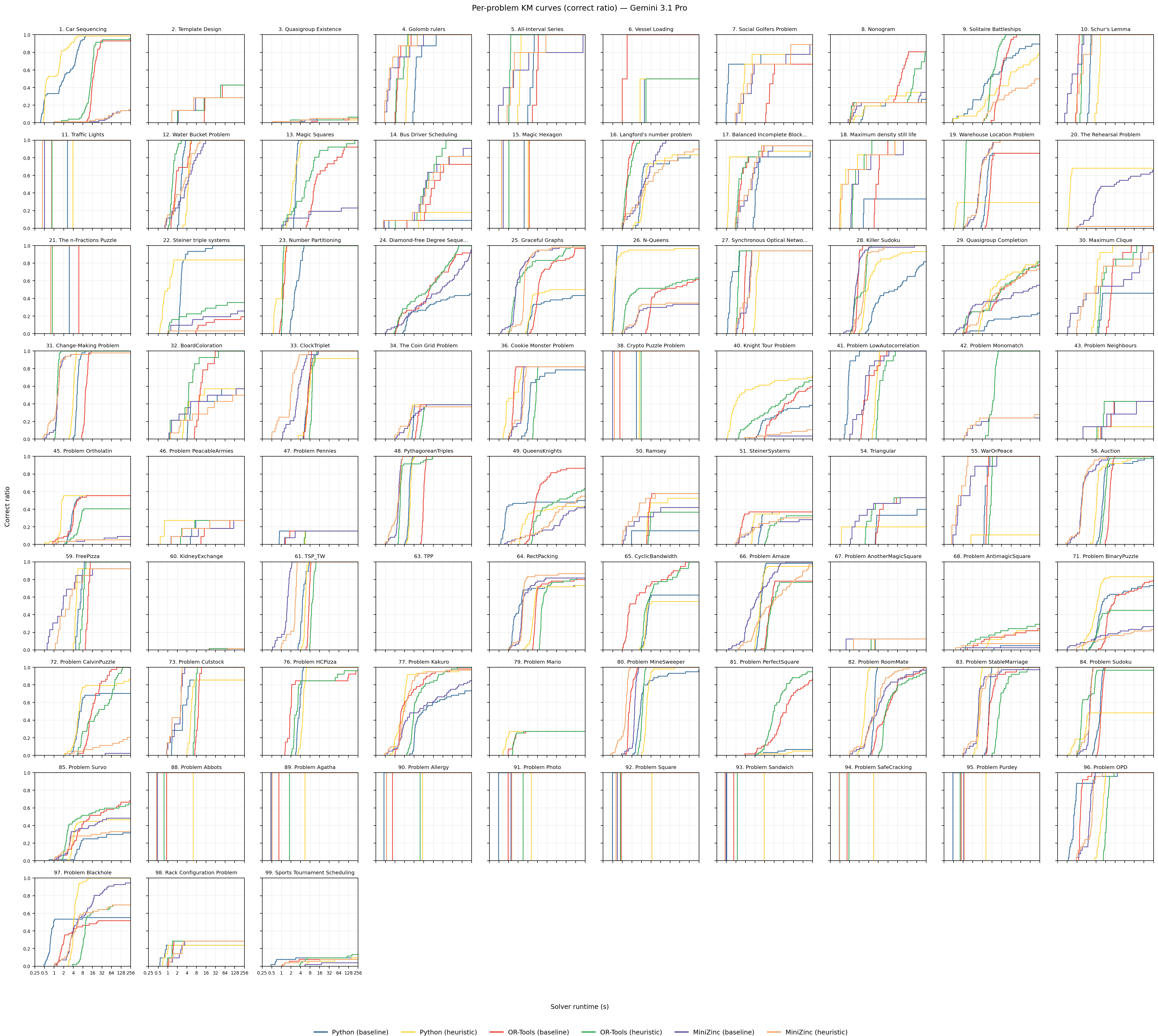}
\caption{Per-problem cumulative runtime curves (correct ratio) for
\textsc{Gemini 3.1 Pro}.}
\label{fig:per-problem-gemini}
\end{figure}

Figs.~\ref{fig:per-problem-codex}, \ref{fig:per-problem-deepseek},
and \ref{fig:per-problem-gemini} report the per-problem cumulative runtime curves
for each LLM, with one subplot per benchmark problem and six curves
per subplot ($3$ paradigms $\times$ $2$ prompts, correct ratio only).
The small-multiples view exposes the heterogeneity that the macro
cumulative curves of Fig.~\ref{fig:km} average over: a non-trivial fraction
of problems are solved at correct ratio $1.0$ under every setting,
while a complementary fraction are solved at correct ratio $0$ under
every setting; the settings whose macro behaviour we discuss in
Sec.~\ref{sec:heuristic-trap} therefore differ on the
\emph{intermediate-difficulty} band visible in these per-problem
plots. The construct- and mode-conditioning of
Sec.~\ref{sec:trap-bypattern} identifies the specific subset of
problems on which each (LLM, paradigm) pair regresses; we recommend
the per-problem plots primarily as a visual aid when reading those
tables.

\section{Anatomy of Baseline Silent Failures: The NL-to-Formal-Language Gap}
\label{sec:baseline-modes}

Sec.~\ref{sec:heuristic-trap} explains the \emph{differential}
between \textit{heuristic} and \textit{baseline} prompts. It does not
explain the silent-failure ranking already visible on the
\textit{baseline} prompt itself: even with no heuristic instruction,
every (paradigm, LLM) pair has a non-trivial provided--correct gap,
and that gap concentrates differently across paradigms. This
appendix dissects the baseline gap mechanistically. The framing is
that ``provided but wrong'' on the \textit{baseline} prompt is, in
essentially every case, a failure at the natural-language-to-formal-language
boundary rather than a failure of the underlying solver---CP-SAT
and the MiniZinc flattener are sound, so any verifier-rejected
solution traces to the encoding step. Five mechanisms account for
nearly all of the baseline silent failures: four (\textbf{M1--M4})
appear on every paradigm, and a fifth (\textbf{M5}) is unique to
\paradigm{Python}, where the LLM additionally writes the search
procedure.

\subsection{From Symptom to Cause: Encoding Non-Equivalence}
\label{sec:baseline-equiv}

Let $F_{\text{ref}}(x)$ denote the reference verifier's feasible set
for an instance $x$, and $F_{\text{llm}}(x)$ the feasible set of the
LLM-generated formal model. The verdict the evaluation pipeline
records is determined by the relation between the two sets together
with the solver's choice within $F_{\text{llm}}(x)$. When
$F_{\text{llm}}(x)\!\subsetneq\!F_{\text{ref}}(x)$ (over-constraint),
the LLM's optimum is feasible in the reference and the verdict is
\emph{correct} on a CSP, or \emph{suboptimal} on a COP whenever the
over-constraint forecloses the true optimum; if the LLM's feasible
set is empty, the verdict is \emph{no solution}. When
$F_{\text{llm}}(x)\!\not\subseteq\!F_{\text{ref}}(x)$
(under-constraint or mis-encoding), CP-SAT picks an assignment that
the verifier rejects when it pins the assignment into the reference
model, and the verdict is \emph{UNSAT}.

The empirical signature of the baseline gap is therefore diagnostic:
\emph{UNSAT} dominates whenever encoding is non-equivalent in a
direction that lets the LLM's solver leave the reference's feasible
region; \emph{suboptimal} dominates whenever encoding is
over-constrained but still inside the reference region. For the
\paradigm{Python + OR-Tools} \textit{baseline} setting,
Tab.~\ref{tab:baseline-wrong-bucket} reports the wrong-mass split
across the three LLMs on the shared $N\!=\!2{,}565$ pool: across all
three LLMs, $\geq\!83.8\%$ of the wrong mass is \emph{UNSAT}, and the
small \emph{suboptimal} share is essentially all on COPs where the
encoding tightens but does not exit the feasible region. The same
diagnosis holds on the \paradigm{MiniZinc + OR-Tools}
\textit{baseline} setting; only the absolute provided mass differs
because the MiniZinc surface causes many instances to drop into
\emph{no solution} before the encoding ever reaches the verifier.

\begin{table}[t]
\centering
\caption{\textit{Baseline} wrong-mass split for the
\paradigm{Python + OR-Tools} setting, per LLM, on the shared
$N\!=\!2{,}565$ pool. \emph{UNSAT}: verifier pins the LLM's
assignment into the reference and the model becomes infeasible.
\emph{suboptimal}: COP solution is feasible in the reference but
worse than the optimum. \emph{invalid}: residual cases.}
\label{tab:baseline-wrong-bucket}
\small
\setlength{\tabcolsep}{6pt}
\begin{tabular}{lrrrr}
\toprule
\textbf{LLM} & Total wrong (frac.) & \emph{UNSAT} & \emph{suboptimal} & \emph{invalid} \\
\midrule
\textsc{DeepSeek-V3.2}    & $469$ ($18.3\%$) & $\mathbf{393}\,(83.8\%)$ & $\phantom{0}68\,(14.5\%)$ & $\phantom{00}8\,(1.7\%)$ \\
\textsc{GPT-5.3-Codex}    & $214$ ($\phantom{0}8.3\%$) & $\mathbf{188}\,(87.9\%)$ & $\phantom{0}19\,(\phantom{0}8.9\%)$  & $\phantom{00}7\,(3.3\%)$ \\
\textsc{Gemini 3.1 Pro}   & $256$ ($10.0\%$) & $\mathbf{256}\,(100.0\%)$ & $\phantom{00}0\,(\phantom{0}0.0\%)$  & $\phantom{00}0\,(0.0\%)$ \\
\bottomrule
\end{tabular}
\end{table}

\subsection{Mechanism Inventory (M1--M4)}
\label{sec:baseline-mechanisms}

The four mechanisms below cover $\approx\!90\%$ of the audited
\emph{UNSAT} mass on the \paradigm{Python + OR-Tools}
\textit{baseline} setting (top-12 problems,
Sec.~\ref{sec:baseline-distribution}); they are
NL-to-formal-language mechanisms and apply identically on
\paradigm{MiniZinc + OR-Tools}.

\paragraph{M1. Associative recall: the LLM names the problem before
it reads it.} Frontier LLMs index problems by \emph{type} before
they finish reading the constraint list. ``Travelling Purchaser''
invokes a TSP template; ``Kidney Exchange'' invokes max-weight
matching; ``Mario fuel-bounded path'' invokes shortest-path-with-budget.
Each retrieval is approximately right but carries idiomatic
side-effects (Hamiltonian over all cities, single canonical
successor, MTZ ordering) that the actual problem does not require.
On problem~63 (TPP), \textsc{GPT-5.3-Codex} and \textsc{DeepSeek-V3.2}
both reflexively impose a Hamiltonian circuit over every city---two
independent LLMs converging on the same template:
\begin{lstlisting}[style=pyinline]
# GPT-5.3-Codex, problem 63 (TPP)
arcs = [(i, j, x[(i, j)]) for (i, j) in x]
model.AddCircuit(arcs)
\end{lstlisting}
The TPP description allows the tour to skip non-purchase locations;
the template forecloses that, so any instance whose true optimum
visits fewer cities than the cheapest full Hamiltonian becomes
infeasible in $F_{\text{llm}}$.

\paragraph{M2. Long-spec attention drift: the LLM encodes most of
the constraints, just not all.} CP-SynC-XL problem descriptions list
$4$--$10$ textual constraints, some inline in prose, some in the
input/output schema. Faithful encoding requires an explicit clause
for each. If the LLM transcribes seven of nine and silently skips
two, CP-SAT solves the seven-constraint model to optimality, the
verifier checks all nine, and rejects. This structural
under-constraint is the single biggest source of \emph{UNSAT} on
problems whose specification runs across multiple paragraphs
(Rehearsal, Coin Grid, TSP\_TW, Nonogram). The pipeline cannot
recover from M2: the LLM's model is internally consistent, CP-SAT
reports \texttt{OPTIMAL}, and there is no failure signal to feed
back.

\paragraph{M3. Output-schema ambiguity between LLM and verifier.}
The verifier's evaluation step pins the LLM's claimed values into a
reference model whose decision-variable schema the LLM has to match
exactly. The natural-language description constrains the
\emph{output} schema (keys, types, sizes), but it rarely specifies
how those keys are wired into the reference's variables. The LLM
has to guess: should \texttt{tour} include the depot once or twice?
Is \texttt{successors[i]\!=\!i} the convention for ``unmatched'',
or a forbidden self-loop? Is \texttt{model\_of\_rack[i]\!=\!-1} the
sentinel for ``unused'', or does the verifier expect a missing key?
Each plausible-looking guess can either match the reference
(\emph{correct}) or not (\emph{UNSAT}). Problem~60 (KidneyExchange)
is the cleanest example: all three LLMs return
$\texttt{successors}[v]\!=\!v$ for unmatched pairs, but the
reference verifier rejects self-loops as the encoding of ``no
exchange''. M3 concentrates in problems with explicit
``unused / unmatched / placeholder'' output semantics.

\paragraph{M4. Sentinel-value misreading.} Benchmark inputs use
sentinel values for ``no edge'', ``incompatible'', or ``infinite
cost''---typically $-1$, $0$, or a large positive number. The
natural-language description rarely names the convention explicitly,
and a wrong reading drops or keeps edges against the spec.
\textsc{GPT-5.3-Codex}'s \textit{baseline} Mario solver:
\begin{lstlisting}[style=pyinline]
# GPT-5.3-Codex, problem 79 (Mario)
for i in range(n):
    for j in range(n):
        if i != j and fuels[i][j] > 0:   # reads 0 as "no edge"
            arcs.append((i, j))
\end{lstlisting}
treats $\texttt{fuels}[i][j]\!=\!0$ as ``no edge'', whereas the
dataset uses $0$ as ``edge present, free move''. The LLM has
unilaterally banned every free arc; \emph{UNSAT} on every instance
whose true optimum needs one.

\subsection{Per-Problem Audit and Mechanism Distribution}
\label{sec:baseline-distribution}

We manually audited the top twelve problems by aggregated
\emph{UNSAT} mass on the \paradigm{Python + OR-Tools}
\textit{baseline} setting---these account for $750$ of $837$
instances ($89.6\%$) of the wrong mass across all three LLMs
(Tab.~\ref{tab:baseline-audit}). Aggregating dominant mechanism
across the audited $750$ instances yields
M1 $120$ ($16\%$),
M2 $282$ ($38\%$),
M3 $270$ ($36\%$),
M4 $\phantom{0}78$ ($10\%$); see also the discussion in
Sec.~\ref{sec:exp-outcomes}.

\begin{table}[t]
\centering
\caption{Per-problem audit of \paradigm{Python + OR-Tools}
\textit{baseline} \emph{UNSAT} mass. ``DS / Cd / Gm'' is the
per-instance count of \emph{UNSAT} verdicts on
\textsc{DeepSeek-V3.2}, \textsc{GPT-5.3-Codex}, and
\textsc{Gemini 3.1 Pro}. ``Mech.'' is the dominant assigned
mechanism from Sec.~\ref{sec:baseline-mechanisms}; problem~63 splits
$50\%/50\%$ between M1 (Hamiltonian template) and M2
(purchase-coupling under-constraint).}
\label{tab:baseline-audit}
\small
\setlength{\tabcolsep}{4pt}
\begin{tabular}{rlrrrl}
\toprule
\textbf{pid} & \textbf{Problem} & DS & Cd & Gm & \textbf{Mech.} \\
\midrule
$63$ & TPP (Travelling Purchaser)               & $55$ & $37$ & $60$ & M1+M2 \\
$60$ & KidneyExchange (max-weight cycles)       & $37$ & $36$ & $35$ & M3 \\
$34$ & Coin Grid                                & $25$ & $25$ & $25$ & M2 \\
$20$ & Rehearsal Problem                        & $14$ & $14$ & $44$ & M2 \\
$66$ & Amaze (line-pair routing)                & $56$ & $\phantom{0}0$  & $13$ & M3 \\
$79$ & Mario (fuel-bounded path)                & $\phantom{0}0$  & $29$ & $22$ & M4 \\
$98$ & Rack Configuration                       & $14$ & $15$ & $15$ & M3 \\
$72$ & CalvinPuzzle                             & $44$ & $\phantom{0}0$  & $\phantom{0}0$  & M1 \\
$81$ & PerfectSquare                            & $40$ & $\phantom{0}0$  & $\phantom{0}0$  & M3 \\
$61$ & TSP\_TW                                  & $28$ & $\phantom{0}0$  & $\phantom{0}0$  & M2 \\
$97$ & Blackhole                                & $\phantom{0}0$  & $\phantom{0}0$  & $27$ & M4 \\
$\phantom{0}8$ & Nonogram                       & $26$ & $\phantom{0}0$  & $\phantom{0}0$  & M2 \\
\midrule
\textit{Other 13 small-count problems}          & --   & --   & --   & mixed \\
\bottomrule
\end{tabular}
\end{table}

The audit yields three structural readings.
\textbf{(B1)} M2 and M3 jointly account for $\sim\!74\%$ of the
silent-failure mass, in roughly equal measure; pure problem-template
substitution (M1) is real but smaller in isolation ($16\%$), and
sentinel misreading (M4) is the smallest ($10\%$).
\textbf{(B2)} The same problems fail across all three LLMs:
problems~63, 60, 34, 98 fail on every LLM and contribute
$\geq\!100$ wrong instances each. This shared failure pattern is
strong evidence that the baseline gap reflects a problem-statement
property---ambiguous output schema, long constraint list, an
associative-recall trigger---rather than a per-LLM bug.
\textbf{(B3)} Refinement cannot fix M2: the LLM's model is
internally consistent, CP-SAT reports \texttt{OPTIMAL}, and no
signal flags the missing clause. Sec.~\ref{sec:trap-refinement}
documents that refinement is an error-recovery loop rather than a
performance loop; M2 is the formal reason that loop has a hard
ceiling on baseline correctness.

\subsection{Cross-Paradigm Reading and the \paradigm{Python}-Specific Mechanism (M5)}
\label{sec:baseline-cross-paradigm}

All four mechanisms (M1--M4) are intrinsic to the
NL-to-formal-language step, which is paradigm-agnostic.
\paradigm{MiniZinc + OR-Tools}'s declarative semantics catch
\emph{syntactic} and \emph{type} errors at flatten time, but they do
not catch (i)~the wrong template applied to the right variables (M1),
(ii)~a forgotten constraint clause (M2), (iii)~an output-schema
mismatch between the LLM's emitted Python output formatter and the
verifier's expected schema (M3), or (iv)~the LLM's reading of an
input sentinel (M4). Empirically,
\paradigm{MiniZinc + OR-Tools}'s \emph{conditional} wrong rate
(wrong $\!\mid\!$ provided) is approximately equal to
\paradigm{Python + OR-Tools}'s on the strongest LLMs
(\textsc{GPT-5.3-Codex} \textit{baseline}: $12.5\%$ MiniZinc vs.\
$10.6\%$ OR-Tools); the absolute difference is overwhelmingly a
translation-fluency effect on \paradigm{MiniZinc + OR-Tools}'s
\emph{no solution} bin, not a property of the declarative surface.

\paragraph{M5. Search-procedure bugs (\paradigm{Python}-only).}
\paradigm{Python} carries M1--M4 plus a fifth mechanism: the LLM
writes both the constraint encoding \emph{and} the search procedure,
so wrong mass also includes (i)~unsound pruning rules,
(ii)~off-by-one base cases in DP / DFS, (iii)~premature termination
of incomplete searches, and (iv)~tie-breakers that drop optimal
extensions. M5 takes two shapes: \textbf{M5a} (COP local-optimum
settle), where an incomplete search returns a feasible but
non-optimal assignment that the verifier rejects on optimality
(\emph{suboptimal}); and \textbf{M5b} (junk fallback), where on
hitting an internal node / iteration / time budget, the artefact
returns a schema-conformant but trivially infeasible assignment
(e.g.\ all squares stacked at the corner) that the verifier rejects
on feasibility (\emph{UNSAT}). Both shapes therefore land in the
dominant \emph{UNSAT} or \emph{suboptimal} bins of
Fig.~\ref{fig:outcomes}, not in \emph{invalid}: the verdict bucket
alone is not sufficient to separate M5 from M1--M4, and the
separation has to be done at the problem and code level.

\subsection{\paradigm{Python} \textit{Baseline} Audit: How M1--M4 Propagate and Where M5 Dominates}
\label{sec:baseline-python-audit}

\paradigm{Python}'s \textit{baseline} wrong-mass split per LLM
(Tab.~\ref{tab:python-bucket-split}) shows that \emph{UNSAT}
dominates the wrong mass exactly as on \paradigm{Python + OR-Tools}
---between $82\%$ and $86\%$ on every LLM. The contrast with
\paradigm{Python + OR-Tools} is therefore not the bucket distribution
but the \emph{absolute} size of the \emph{UNSAT} bucket:
\paradigm{Python} carries $24.6\%$ \emph{UNSAT} for
\textsc{DeepSeek-V3.2} vs.\ $15.3\%$ on \paradigm{Python + OR-Tools},
and the gap shrinks to roughly zero on \textsc{Gemini 3.1 Pro}
(which abstains rather than guesses).

\begin{table}[t]
\centering
\caption{\paradigm{Python} \textit{baseline} wrong-mass split per
LLM, on the shared $N\!=\!2{,}565$ pool. The \emph{UNSAT} bucket
dominates exactly as on \paradigm{Python + OR-Tools}
(Tab.~\ref{tab:baseline-wrong-bucket}); the bucket distribution is
therefore paradigm-agnostic, and the difference between
\paradigm{Python} and \paradigm{Python + OR-Tools} is the
\emph{absolute} size of the \emph{UNSAT} bucket.}
\label{tab:python-bucket-split}
\small
\setlength{\tabcolsep}{6pt}
\begin{tabular}{lrrrr}
\toprule
\textbf{LLM} & Total wrong (frac.) & \emph{UNSAT} & \emph{suboptimal} & \emph{invalid} \\
\midrule
\textsc{DeepSeek-V3.2}    & $739$ ($28.8\%$) & $\mathbf{631}\,(85.4\%)$ & $\phantom{0}87\,(11.8\%)$ & $\phantom{0}21\,(2.8\%)$ \\
\textsc{GPT-5.3-Codex}    & $377$ ($14.7\%$) & $\mathbf{311}\,(82.5\%)$ & $\phantom{0}55\,(14.6\%)$ & $\phantom{0}11\,(2.9\%)$ \\
\textsc{Gemini 3.1 Pro}   & $262$ ($10.2\%$) & $\mathbf{226}\,(86.3\%)$ & $\phantom{0}28\,(10.7\%)$ & $\phantom{00}8\,(3.1\%)$ \\
\bottomrule
\end{tabular}
\end{table}

To separate the two failure mechanisms, we cross-reference the top
\paradigm{Python}-\textit{baseline} failures against the
\paradigm{Python + OR-Tools}-\textit{baseline} failures on the same
problems. A problem that fails on \emph{both} paradigms is
diagnostic of M1--M4 (the LLM made the same NL-to-formal mistake on
both surfaces); a problem that fails only on \paradigm{Python} is
diagnostic of M5 (the encoding is paradigm-agnostically right, but
the hand-rolled algorithm is buggy). Tab.~\ref{tab:python-audit}
reports the top problems by aggregated wrong mass across all three
LLMs.

\begin{table}[t]
\centering
\caption{Per-problem audit of \paradigm{Python} \textit{baseline}
wrong mass. ``DS / Cd / Gm'' is the per-instance wrong count on
\textsc{DeepSeek-V3.2}, \textsc{GPT-5.3-Codex}, and
\textsc{Gemini 3.1 Pro}. ``ORT?'' marks problems that also appear in
the \paradigm{Python + OR-Tools} \textit{baseline} audit
(Tab.~\ref{tab:baseline-audit}); these inherit the M1--M4 mechanism.
``Mech.'' is the dominant assigned mechanism for the
\paradigm{Python} row: M1--M4 if the same problem fails on
\paradigm{Python + OR-Tools} (the encoding bug propagates), M5a /
M5b if \paradigm{Python}-specific.}
\label{tab:python-audit}
\small
\setlength{\tabcolsep}{4pt}
\begin{tabular}{rlrrrlc}
\toprule
\textbf{pid} & \textbf{Problem} & DS & Cd & Gm & \textbf{ORT?} & \textbf{Mech.} \\
\midrule
$63$ & TPP                          & $60$ & $60$ & $54$ & yes (top) & M1+M2 \\
$81$ & PerfectSquare                & $60$ & $60$ & $\phantom{0}0$  & no & \textbf{M5b} \\
$60$ & KidneyExchange               & $36$ & $35$ & $32$ & yes (top) & M3 \\
$65$ & CyclicBandwidth (COP)        & $36$ & $32$ & $15$ & partial (DS only) & \textbf{M5a} \\
$79$ & Mario                        & $10$ & $40$ & $21$ & yes & M4 \\
$20$ & Rehearsal                    & $14$ & $14$ & $44$ & yes & M2 \\
$25$ & Graceful Graphs              & $45$ & $\phantom{0}0$  & $19$ & partial (DS, Gm) & \textbf{M5b} \\
$68$ & AntiMagicSquare              & $26$ & $38$ & $\phantom{0}0$  & partial (Cd) & \textbf{M5b} \\
$82$ & RoomMate                     & $58$ & $\phantom{0}0$  & $\phantom{0}0$  & no & \textbf{M5b} \\
$49$ & QueensKnights                & $52$ & $\phantom{0}0$  & $\phantom{0}0$  & no & \textbf{M5b} \\
$77$ & Kakuro                       & $48$ & $\phantom{0}0$  & $\phantom{0}0$  & no & \textbf{M5b} \\
$98$ & Rack Configuration           & $\phantom{0}3$  & $15$ & $14$ & yes & M3 \\
$\phantom{0}8$ & Nonogram           & $26$ & $\phantom{0}0$  & $\phantom{0}5$  & yes & M2 \\
\bottomrule
\end{tabular}
\end{table}

The audit yields two readings.
\textbf{(B4) M1--M4 propagate.} The top
\paradigm{Python + OR-Tools} \textit{baseline} failures
(Tab.~\ref{tab:baseline-audit})---p63, p60, p79, p20, p34---all
reappear among the top \paradigm{Python} failures with comparable
per-LLM counts: the TPP / Hamiltonian template, the KidneyExchange
self-loop sentinel, the Mario zero-fuel sentinel, and the
Rehearsal long-spec attention drift are paradigm-agnostic.
Approximately $170$--$190$ instances of \paradigm{Python} wrong
mass per LLM are accounted for by these shared problems; the rest
is M5.
\textbf{(B5) M5b dominates the \paradigm{Python}-only mass.} The
\paradigm{Python}-only failures (problems where \paradigm{Python}
is wrong but \paradigm{Python + OR-Tools} is correct) concentrate on
packing / matching / cycle-finding problems whose NL specification
is short and unambiguous (so neither M2 nor M3 applies) but whose
intrinsic difficulty exceeds the LLM's hand-rolled search budget.
On these problems the LLM consistently emits a node- or
iteration-bounded greedy / DFS / DP and falls back to a junk
schema-conformant assignment when the budget runs out (p81
PerfectSquare, p82 RoomMate, p49 QueensKnights, p3 Quasigroup
Existence, p77 Kakuro---collectively $\sim\!330$ M5b instances).
M5a (\emph{suboptimal}) is small in absolute terms ($170$ total
instances) but concentrated on a handful of COPs where the LLM's
incomplete search settles on the first feasible solution, most
prominently p65 CyclicBandwidth.

Two M5b code patterns typify the failure shape.
\textsc{GPT-5.3-Codex}'s \textit{baseline} on p81 (PerfectSquare)
caps DFS at a node budget and \emph{fabricates} a placement when
the budget runs out:
\begin{lstlisting}[style=pyinline]
# GPT-5.3-Codex, problem 81 (PerfectSquare, CSP) -- M5b
if n > 34 or size > 90:
    return {"placements": [[0, 0] for _ in range(n)]}
node_budget = 200000 if n <= 26 else 90000
# DFS that returns a schema-conformant but infeasible
# placement when budget is exhausted.
\end{lstlisting}
\textsc{DeepSeek-V3.2}'s \textit{baseline} on p49 (QueensKnights)
caps DFS at $100{,}000$ nodes, tries $20$ random starts, and on
failure literally falls back to a row-major enumeration with the
explicit comment ``hope for the best'':
\begin{lstlisting}[style=pyinline]
# DeepSeek-V3.2, problem 49 (QueensKnights, CSP) -- M5b
def dfs(start, current, visited, path):
    counter[0] += 1
    if counter[0] > 100000: return None
    # ...
# After 20 random starts fail:
# Create a simple path and hope for the best
path = [(i // n, i % n) for i in range(min(m, n * n))]
\end{lstlisting}
The pipeline records \emph{UNSAT} on every instance whose true
cycle is not on the row-major path.

\subsection{Estimated Mechanism Share on the \paradigm{Python} \textit{Baseline} Setting}
\label{sec:baseline-python-share}

Combining Tabs.~\ref{tab:python-bucket-split} and
\ref{tab:python-audit} via cross-paradigm overlap (problems that
fail on both \paradigm{Python + OR-Tools} and \paradigm{Python}
attribute their \paradigm{Python} wrong mass to M1--M4; problems
that fail only on \paradigm{Python} attribute their wrong mass to
M5) gives the approximate decomposition in
Tab.~\ref{tab:python-mechanism-share}.

\begin{table}[t]
\centering
\caption{Approximate decomposition of \paradigm{Python}
\textit{baseline} wrong mass per LLM into M1--M4
(NL-to-formal mistakes that also fail on \paradigm{Python + OR-Tools})
and M5 (\paradigm{Python}-only search-procedure bugs). The M1--M4
share is estimated by intersecting the \paradigm{Python}-failing
problems with the \paradigm{Python + OR-Tools}-failing problems on
the same LLM; the residual is treated as M5.}
\label{tab:python-mechanism-share}
\small
\setlength{\tabcolsep}{6pt}
\begin{tabular}{lrrr}
\toprule
\textbf{LLM} & \paradigm{Python} wrong & M1--M4 share & M5 share \\
\midrule
\textsc{DeepSeek-V3.2}    & $739$ & $\sim\!360$ ($\sim\!49\%$) & $\sim\!379$ ($\sim\!51\%$) \\
\textsc{GPT-5.3-Codex}    & $377$ & $\sim\!140$ ($\sim\!37\%$) & $\sim\!237$ ($\sim\!63\%$) \\
\textsc{Gemini 3.1 Pro}   & $262$ & $\sim\!187$ ($\sim\!71\%$) & $\sim\!75$  ($\sim\!29\%$) \\
\bottomrule
\end{tabular}
\end{table}

The decomposition refines the design implication: M5 is not a
side-channel. On \textsc{GPT-5.3-Codex} it accounts for the majority
of the \paradigm{Python} \textit{baseline} wrong mass; on
\textsc{DeepSeek-V3.2} M1--M4 and M5 are roughly equal contributors;
and on \textsc{Gemini 3.1 Pro} M1--M4 dominate
(\textsc{Gemini 3.1 Pro} is unusually willing to abstain rather than
fabricate, which suppresses M5b but not M1--M4). The cross-LLM
divergence is the same divergence visible in
Sec.~\ref{sec:trap-cross-llm}: the LLM's algorithmic
\emph{taste}---specifically, its willingness to defer to verified
machinery rather than substitute its own---is the dominant
determinant of both the heuristic-prompt safety profile and the
baseline silent-failure profile on \paradigm{Python}.

\section{Heuristic Trap: Detailed Mode Mechanisms and Code Audits}
\label{app:trap-modes-details}

This appendix expands the brief discussion of the six failure modes
in Sec.~\ref{sec:trap-modes}. The per-(mode, LLM) outcome split is
reported as Tab.~\ref{tab:failure-mode-outcome} in
Sec.~\ref{sec:trap-modes}; here we give one source-level code extract
per mode. Each mode is associated with an operational
\emph{detector}---a code-level signature that lets us mark every
(\textit{LLM}, problem) pair as exhibiting the mode or not (full
detector definitions in App.~\ref{app:mode-detectors}). These
detectors are reproducible proxies for recurring code patterns, not
formal proofs that the pattern caused every downstream failure. We
therefore use them for incidence and conditioning, and validate the
mechanistic interpretation through the paired baseline/heuristic diffs
and representative manual code audits below. The detectors are
intentionally conservative (false-negatives preferred to
false-positives), so the incidence numbers should be read as lower
bounds on the corresponding pattern families.

\paragraph{Mode A (\paradigm{Python + OR-Tools}): ``tight'' bounds the LLM never proves
[$13.5\%$ overall; \textsc{DeepSeek-V3.2} $-6.7$\,pp].}
The \textit{heuristic} prompt explicitly invites the model to
``propagate tight objective bounds''. In practice the LLM
\emph{invents} bounds that look plausible but are not theorems. The
detector flags any heuristic-prompt \paradigm{Python + OR-Tools}
solver that introduces a binary-search wrapper on the objective,
declares an auxiliary multiplicative variable with a hand-derived
domain, or inserts a comment asserting a ``dominance'' / ``valid
inequality'' / ``tight bound''
(Tab.~\ref{tab:failure-mode-outcome}, row~A). In problem~2
(multi-template pressing, COP) under \textsc{GPT-5.3-Codex}, the
\textit{baseline} keeps optimisation native; the \textit{heuristic}
version replaces \texttt{Minimize} with a binary search over total
pressings $P$ and at each step rebuilds a feasibility model whose
auxiliary product variables $z_{k,i}\!=\!x_{k,i}\!\cdot\!p_k$ are
constrained by a hand-derived per-element upper bound:
\begin{lstlisting}[style=pyinline]
excess = S * P - D
zub = max(0, min(S * P, d[i] + excess))   # claimed dominance bound
zik = model.NewIntVar(0, zub, f"z_{k}_{i}")
model.AddMultiplicationEquality(zik, [x[k][i], p[k]])
\end{lstlisting}
The bound depends on a distributional argument the LLM never
states---that no optimal solution over-covers any single variation by
more than the global slack---which fails on instances where one
variation must absorb most of the slack to keep other rows feasible.
The feasibility model becomes infeasible at the true optimum, the
binary search overshoots, and correctness drops from $1.00$ to
$0.43$.

\paragraph{Mode B (\paradigm{Python + OR-Tools}): semantics-preserving rewrites with weaker
propagation [$4.0\%$ overall; $33\%$ within-mode regression rate].}
A second class of regressions arises from rewrites that are
\emph{mathematically equivalent} to the \textit{baseline} but cripple
CP-SAT propagation. The detector triggers when the heuristic-prompt
solver \emph{newly} introduces \texttt{AddCircuit} or a big-$M$ /
MTZ encoding that the \textit{baseline} did not contain. The mode is
rare but high-risk: $33.3\%$ of within-mode problems regress and
$21.0\%$ of common-correct instances are slower than \textit{baseline}
(Tab.~\ref{tab:failure-mode-outcome}, row~B). In problem~49 (queens
+ knight cycle, CSP) the \textit{heuristic} solver replaces
\textit{baseline} cycle-position variables with a directed circuit
over all $n^2$ board cells with self-loops on unselected cells:
\begin{lstlisting}[style=pyinline]
self_loop = [k_model.NewBoolVar(f"self_{i}") for i in range(N)]   # N = n*n
arcs = [(i, i, self_loop[i]) for i in range(N)] + edge_arcs
k_model.Add(sum(self_loop) == N - m)
k_model.AddCircuit(arcs)
\end{lstlisting}
Valid, but with $\Theta(n^2)$ Boolean variables and a global circuit
over them: the \textit{heuristic} version exhausts the $256$\,s
budget at instance $38$ where the \textit{baseline} solves up to $52$
in $\!\leq\!14$\,s. The same pattern appears in problem~79 (Mario
fuel-bounded path), where the LLM swaps the idiomatic reified MTZ
\texttt{model.Add(u[j]>=u[i]+1).OnlyEnforceIf(x)} for a textbook
big-$M$ encoding $u[j]\!\geq\!u[i]\!+\!1\!-\!M(1\!-\!x)$ that is
strictly weaker for CP-SAT.

\paragraph{Mode C (\paradigm{Python + OR-Tools}): hand-computed
seeding under the heuristic prompt [$19.5\%$ overall;
$9$--$26\%$ per LLM].}
The third \paradigm{Python + OR-Tools} pattern adds a CP-SAT
hand-computed candidate before invoking the solver. The safe version
is advisory warm-starting (\texttt{AddHint}); the risky version turns
the same candidate into hard equalities
(\texttt{model.Add(... == k)}). We group them because both originate
from the same LLM behavior---constructing a separate greedy or
domain-specific candidate and feeding it to CP-SAT---but we separate
their implications in the interpretation: hints can slow search but
do not change the feasible set, while hard fixes can remove feasible
or optimal assignments if the candidate is wrong. The candidate may
be produced by a named helper (\texttt{def warnsdorff} /
\texttt{def greedy} / \texttt{def construct\_*}) or by an inlined
``\texttt{initial\_*}'' computation inside
\texttt{solve\_instance}. We flag the mode whenever such a seeding
mechanism appears in the heuristic preset and is absent from the
baseline. Under this behavioural detector the mode is the most
popular \paradigm{Python + OR-Tools} heuristic intervention in our
audit ($26.3\%$ of \textsc{GPT-5.3-Codex}, $9.1\%$ of
\textsc{Gemini 3.1 Pro}, $23.2\%$ of \textsc{DeepSeek-V3.2};
Tab.~\ref{tab:failure-mode-outcome}, row~C); the \emph{within-mode}
mean correctness change is mildly positive on every LLM ($+0.009$
to $+0.061$) and the median per-problem speed-up sits at
$0.96$--$1.02\times$. This near-neutral aggregate is mostly explained
by advisory hints; the correctness risk concentrates in the smaller
subset where the seed is enforced as a hard constraint.

The two extremes of within-mode behaviour come from whether the
hand-coded candidate is sound for the problem class and whether CP-SAT
is allowed to ignore it. In problem~40
(closed knight's tour, CSP) \textsc{GPT-5.3-Codex} implements
Warnsdorff's rule with multi-restart, then \emph{constrains} every
CP-SAT arc to follow the discovered cycle:
\begin{lstlisting}[style=pyinline]
def heuristic_cycle():
    for attempt in range(max_attempts):
        # ... Warnsdorff: pick next vertex with min remaining degree ...
        return path                                                 # full closed tour
cycle_path = heuristic_cycle()
fixed_next = {cycle_path[i]: cycle_path[(i+1) % N] for i in range(N)}
model.AddCircuit(arcs)
for u in range(N):                                                  # hard equality
    model.Add(arc_var[(u, fixed_next[u])] == 1)
\end{lstlisting}
Here the discovered tour is a valid witness, so the hard equality is
safe on instances where the helper succeeds and accuracy jumps from
$0.067$ to $0.567$. The opposite case appears when the LLM seeds with
a suboptimal greedy guess but uses only a soft hint: the hint can
anchor search around a worse region of the polytope, yet correctness
survives because CP-SAT continues exploring beyond the seed. This is
why the within-mode mean $\Delta$\,acc stays near zero even when
individual seeds are unhelpful. The
\textsc{DeepSeek-V3.2} mode-incidence in particular is dominated
by inlined warm-starts (no \texttt{def} factor-out), confirming
that the original named-helper signature was Codex-idiomatic
rather than mechanism-defining.

\paragraph{Mode D (\paradigm{Python}): silent loss of completeness
[$16.2\%$ overall; \textsc{DeepSeek-V3.2} $23.2\%$ incidence with $-3.7$\,pp expected].}
On \paradigm{Python} the LLM is responsible for the entire algorithm,
and the \textit{heuristic} prompt induces it to substitute a
\emph{complete} search procedure for an \emph{incomplete} one without
flagging the loss. The detector triggers on any heuristic-prompt
\paradigm{Python} solver that introduces local-search /
simulated-annealing / multi-restart / iteration-cap machinery while
the \textit{baseline} \emph{either} contained a complete-search
construct (\texttt{lru\_cache} / \texttt{permutations} / DP table)
that was \emph{dropped} \emph{or} did not contain those incomplete
constructs in the first place (i.e.\ they were added by the
\textit{heuristic} prompt). The mode is moderately common ($16.2\%$
overall) and exhibits clear LLM-conditioning: \textsc{GPT-5.3-Codex}
stays mildly positive ($+1.8$\,pp), \textsc{Gemini 3.1 Pro} positive
($+5.5$\,pp), \textsc{DeepSeek-V3.2} regresses by $3.7$\,pp on
average with $21.7\%$ of within-mode problems regressing
(Tab.~\ref{tab:failure-mode-outcome}, row~D). The clearest case is
problem~68 (anti-magic square) under \textsc{GPT-5.3-Codex}: the
\textit{baseline} keeps an exhaustive permutation branch for the
small case alongside a uniform stochastic local search:
\begin{lstlisting}[style=pyinline]
# baseline (Python)
if n == 3:
    for p in permutations(nums):                  # complete on n <= 3
        if is_valid(compute_sums(p)):
            return {"grid": to_grid(list(p))}
# else: SA with uniform random swap, return best_arr
\end{lstlisting}
The \textit{heuristic} version drops the exhaustive branch entirely
and replaces the uniform random swap with a \emph{biased} move
generator that concentrates $85\%$ of moves on cells participating
in ``bad'' lines---duplicate or extreme line-sums---so as to converge
faster:
\begin{lstlisting}[style=pyinline]
# heuristic (Python)
bad_lines = [ln for ln, x in enumerate(ls) if freq[x] > 1]
if mx - mn + 1 > m:
    bad_lines += [ln for ln, x in enumerate(ls) if x in (mn, mx)]
bad_cells = [p for ln in bad_lines for p in line_cells[ln]]
# inside the move loop:
i = bad_cells[rng.randrange(len(bad_cells))] if bad_cells and rng.random() < 0.85 \
    else rng.randrange(N)
\end{lstlisting}
The biased generator descends to a local optimum faster, but it can
also keep the search there: correctness drops from
$0.826$ to $0.087$, the largest single-problem regression in our
entire benchmark. Globally, memoisation usage \emph{drops} from
\textsc{GPT-5.3-Codex}'s $6$ to $2$ files under the
\textit{heuristic} prompt and multi-restart usage \emph{appears} from
$0$ to $5$ files (Tab.~\ref{tab:patterns-per-backend}). The pattern
is amplified in \textsc{DeepSeek-V3.2}: $18.2\%$ of problems use
multi-restart and $37.4\%$ use randomised choice---the within-pattern
regression rate of iteration-cap truncation reaches $41.7\%$
(Tab.~\ref{tab:trap-by-pattern}).

\paragraph{Mode E (\paradigm{MiniZinc + OR-Tools}): COP$\to$SAT collapse via assumed optimum
[$0.7\%$ overall; expected $\boldsymbol{-33.4}$\,\textbf{pp}].}
Specific to \paradigm{MiniZinc + OR-Tools}, the \textit{heuristic}
prompt occasionally elicits a hidden
\emph{optimization-to-satisfaction} substitution: the LLM assumes the
optimum equals a known lower bound supplied as input and removes the
objective. The detector triggers when the \textit{baseline} used
\texttt{solve minimize} / \texttt{maximize} but the \textit{heuristic}
version uses \texttt{solve satisfy}, or when the \textit{heuristic}
declares \texttt{var X..X: $v$} (a single-value domain bound to an
input parameter). The mode is rare overall ($2$ of $98$
\textsc{GPT-5.3-Codex} problems, none of \textsc{Gemini 3.1 Pro}'s or
\textsc{DeepSeek-V3.2}'s match the strict signature) but its expected
effect within the mode is $-33.4$\,pp---the largest mean regression of
any mode (Tab.~\ref{tab:failure-mode-outcome}, row~E). Problem~14
(bus-driver shift selection, $\min\#\text{shifts}$) under
\textsc{GPT-5.3-Codex} illustrates the failure. The
\textit{baseline} keeps the COP intact:
\begin{lstlisting}[style=pyinline]
var min_num_shifts..num_shifts: total_shifts;
constraint total_shifts = sum(s in SHIFTS)(x[s]);
solve minimize total_shifts;
\end{lstlisting}
The \textit{heuristic} version's final refinement freezes
\texttt{total\_shifts} to the data-supplied lower bound and discards
the objective:
\begin{lstlisting}[style=pyinline]
% "Known optimal cardinality: turn optimization into satisfaction"
var min_num_shifts..min_num_shifts: total_shifts;
constraint total_shifts = min_num_shifts;
constraint sum(s in SHIFTS)(x[s]) = total_shifts;
solve :: int_search(x, first_fail, indomain_max, complete) satisfy;
\end{lstlisting}
\texttt{min\_num\_shifts} is the input's \emph{lower bound}, not the
proven minimum: when the bound is not tight (most instances) the SAT
problem is infeasible and the evaluation pipeline records \emph{UNSAT}.
Accuracy $0.75\to0.083$. Mode~E is rare in absolute terms but has
the largest within-mode mean regression in our audit; it is
concentrated in COPs whose lower bound is a problem parameter the LLM
can mistake for an optimum.

\paragraph{Mode F (\paradigm{MiniZinc + OR-Tools}): redundant-machinery blow-ups
[$29.1\%$ overall; the empirical kernel of \paradigm{MiniZinc + OR-Tools}'s bimodal outcome].}
A complementary \paradigm{MiniZinc + OR-Tools} failure is symmetric
to \paradigm{Python + OR-Tools} Mode~B but operates at the
\emph{model} level: the LLM augments a clean model with voluminous
``stronger'' machinery and the resulting model can either help
propagation or overwhelm flattening. The detector triggers when
(a)~the \textit{heuristic} file is at least $1.5\times$ the
\textit{baseline} file's line count \emph{and} introduces
direction-tagged auxiliary arrays
(\texttt{pred\_up}/\texttt{down}/\texttt{left}/\texttt{right}),
distance fields, or ``redundant''/``implied'' commentary, or
(b)~when two of those auxiliary signatures co-occur. Mode~F is by
far the most common mode in our benchmark---incidence $29.1\%$
overall, $39.8\%$ for \textsc{GPT-5.3-Codex}, $37.5\%$ for
\textsc{DeepSeek-V3.2} (Tab.~\ref{tab:failure-mode-outcome}, row~F)
---and it is the empirical kernel of
\paradigm{MiniZinc + OR-Tools}'s bimodal outcome: median speed-up
$1.08$--$1.17\times$, $25\%$ of within-mode problems improve by
$\geq\!5$\,pp, and $22.5\%$ regress by $\geq\!5$\,pp. The mode
therefore accounts for both the genuine
\paradigm{MiniZinc + OR-Tools} speed-ups and the heaviest tail of
\paradigm{MiniZinc + OR-Tools} regressions. Problem~66 (Numberlink,
CSP) under \textsc{GPT-5.3-Codex} grows from a $40$-line
\textit{baseline} to a $156$-line \textit{heuristic} version. The
injected machinery is a per-path distance field plus four
direction-specific predecessor booleans plus a checkerboard parity
invariant:
\begin{lstlisting}[style=pyinline]
% added by the heuristic prompt: O(|P|.R.C) auxiliaries
array[P, R, C] of var 0..N: d;                     % distance from source per path
array[P, R, C] of var bool: pred_up;
array[P, R, C] of var bool: pred_down;
array[P, R, C] of var bool: pred_left;
array[P, R, C] of var bool: pred_right;

% "Useful implied constraints" -- in fact a Manhattan-distance lower bound
constraint forall(p in P)(
  d[p, tr[p], tc[p]] >= abs(tr[p]-sr[p]) + abs(tc[p]-sc[p]) + 1
);
% Checkerboard parity (claimed redundant, actually heavy to flatten)
constraint forall(p in P, r in R, c in C)(
  x[p,r,c] -> (d[p,r,c] mod 2 = parity[p,r,c])
);
\end{lstlisting}
Each comment annotates the construction as ``useful implied
constraint'' or ``stronger propagation'', but the flattened model is
too large for CP-SAT to load within the $256$\,s budget on $4$ of $6$
instances. Accuracy drops $1.00\to0.00$, the worst
\paradigm{MiniZinc + OR-Tools} regression for \textsc{GPT-5.3-Codex}.
The same syntactic pattern---adding direction-tagged auxiliary arrays
and ``parity'' constraints to a working channelling model---also
drives the regressions on problems $44$ and $62$.

\paragraph{Mode-detector definitions (operational, reproducible).}
\label{app:mode-detectors}
For completeness we record the exact signatures used to classify
each heuristic-prompt solver into the modes of
Tab.~\ref{tab:failure-mode-outcome}. A solver is flagged for a mode
if \emph{any} disjunct in the corresponding clause is satisfied. The
signatures are designed for reproducibility and high precision, not
exhaustive recall: they identify visible instances of each pattern
family, after which the regression analysis compares the marked
heuristic solver to its paired baseline on the same problem and LLM.
The code examples above illustrate the manual audit used to interpret
the flagged families; unflagged solvers may still contain related
mistakes in forms the detectors do not cover.
\begin{itemize}\itemsep1pt
    \item \textbf{Mode A} (\paradigm{Python + OR-Tools}): heuristic
    source contains \emph{any} of (i)~a \texttt{\#}~comment matching
    the regex
    \texttt{(dominance$|$tight\_bound$|$tightening$|$valid\_inequality
    $|$upper\_bound$|$lower\_bound$|$implied\_bound)},
    (ii)~a binary-search-on-objective wrapper of the form
    \texttt{lo, hi = ...; while lo < hi: ...}, or (iii)~a newly
    introduced \texttt{AddMultiplicationEquality} (absent in the
    \textit{baseline}).
    \item \textbf{Mode B} (\paradigm{Python + OR-Tools}): heuristic
    source \emph{newly} contains \texttt{AddCircuit} or a big-$M$
    pattern (regex
    \texttt{$-$M$\backslash$s$\ast\backslash$$\ast$$\backslash$s$\ast\backslash$($\backslash$s$\ast$1$\backslash$s$\ast$$-$}
    or the keyword \texttt{MTZ}) absent from the \textit{baseline}.
    \item \textbf{Mode C} (\paradigm{Python + OR-Tools}): heuristic
    source contains an \texttt{AddHint} or a hard equality of a
    CP-SAT variable to a constant that is tied to a hand-computed
    candidate, AND \emph{either} (i)~a Python helper named
    \texttt{greedy}, \texttt{warnsdorff},
    \texttt{nearest\_neighbour}, \texttt{construct\_*},
    \texttt{heuristic\_*}, etc., \emph{or}
    (ii)~the same seeding mechanism is absent from the
    \textit{baseline}.  We record whether the seed is advisory
    (\texttt{AddHint}) or enforced (hard equality) for interpretation;
    the table groups them because both are generated by the same
    candidate-construction behavior. Clause~(ii) captures inlined
    warm-starts where the LLM emits the heuristic directly inside
    \texttt{solve\_instance} rather than factoring it into a separately
    named helper.
    \item \textbf{Mode D} (\paradigm{Python}): heuristic source
    contains a local-search / SA / multi-restart / iteration-cap
    construct, AND \emph{either} the \textit{baseline} contained a
    complete-search construct (\texttt{lru\_cache} /
    \texttt{permutations} / DP table) that has been dropped, OR the
    \textit{baseline} did not contain those incomplete-search
    constructs in the first place.
    \item \textbf{Mode E} (\paradigm{MiniZinc + OR-Tools}): the
    \textit{baseline} used \texttt{solve minimize} or
    \texttt{solve maximize} but the \textit{heuristic} uses
    \texttt{solve satisfy}, OR the \textit{heuristic} declares
    \texttt{var X..X: $v$} (single-value domain bound to a parameter).
    \item \textbf{Mode F} (\paradigm{MiniZinc + OR-Tools}): heuristic
    file is at least $1.5\times$ the \textit{baseline} file's line
    count AND introduces a direction-tagged auxiliary array
    (\texttt{pred\_up}/\texttt{down}/\texttt{left}/\texttt{right} or
    similar), or a \texttt{distance} field, or a comment marked
    ``redundant'' / ``implied'' / ``stronger propagation'' /
    ``channelling'' / ``prefix sum''; OR two of those auxiliary
    signatures co-occur regardless of size.
\end{itemize}
The full detector script is released with the benchmark
(\texttt{analyze\_failure\_modes.py}); the per-problem mode labels
are in \texttt{per\_problem\_modes.csv} alongside every accuracy and
runtime measurement.

\section{Model-Specific Heuristic Choices and Safety}
\label{sec:trap-cross-llm}

The aggregate picture of Sec.~\ref{sec:trap-aggregate} shows
\textsc{DeepSeek-V3.2} as the clearest victim of the heuristic trap;
the per-pattern data of Tab.~\ref{tab:trap-by-pattern} explain
\emph{which} mistakes drive its loss. On \paradigm{Python + OR-Tools}
every one of \textsc{DeepSeek-V3.2}'s heavily-used constructs has a
within-pattern regression rate between $26\%$ and $35\%$:
\texttt{AddDecisionStrategy} ($35\%$), \texttt{OnlyEnforceIf}
($31\%$), \texttt{num\_search\_workers} ($27\%$), symmetry-breaking
($27\%$), variable-fixing ($30\%$). The constructs themselves are
correct; \textsc{DeepSeek-V3.2}'s choice of \emph{scope} for them is
not---it pins variables to incorrect values, asserts symmetries that
break the constraint structure, and imposes value orderings that
foreclose feasibility. On \paradigm{MiniZinc + OR-Tools} the same
diagnosis holds: \texttt{first\_fail} regresses on $25\%$ of
\textsc{DeepSeek-V3.2}'s solvers because the chosen variable scope
contains channelling auxiliaries whose \texttt{first\_fail} ordering
forecloses the model. Crucially, $32.0\%$ of \emph{redundant
constraints} in \textsc{DeepSeek-V3.2}'s \paradigm{MiniZinc + OR-Tools}
\textit{heuristic} preset are themselves correctness-breaking
(Tab.~\ref{tab:trap-by-pattern}): the LLM fails to verify that
``redundant'' is, in fact, an implication.

\textsc{Gemini 3.1 Pro} sits at the threshold. Its \paradigm{Python}
heuristics gain $+2.2$\,pp---driven mostly by stronger DFS skeletons
and randomised restarts that happen to suit the problems for which
\textsc{Gemini 3.1 Pro} has weaker default code paths---while its
\paradigm{Python + OR-Tools} and \paradigm{MiniZinc + OR-Tools}
heuristics lose $0.6$--$1.1$\,pp through localised over-specification
(e.g.\ \texttt{global\_lex} regresses on $36\%$ of the problems where
it is used).

\textsc{GPT-5.3-Codex} is the only LLM whose \textit{heuristic}
prompt is net positive on every paradigm, and its safety profile is
largely explained by one behavioral tendency: when in doubt,
\textsc{GPT-5.3-Codex} prefers a native solver hook
(\texttt{AddDecisionStrategy}, \texttt{first\_fail},
\texttt{all\_different}, \texttt{circuit}) over a hand-rolled
algorithmic substitution. On \paradigm{MiniZinc + OR-Tools}
\texttt{circuit} alone delivers a $5.54\times$ median speed-up with
zero regressions; \texttt{all\_different} delivers $1.40\times$ with
$+7.8$\,pp. The trap is therefore not explained by model strength
alone: in this benchmark, an important determinant of whether the
\textit{heuristic} prompt is safe is whether the model expresses the
optimization request through verified solver machinery or through an
unverified algorithmic substitution.

%%%%%%%%%%%%%%%%%%%%%%%%%%%%%%%%%%%%%%%%%%%%%%%%%%%%%%%%%%%%

\newpage

\end{document}